\documentclass[lettersize,journal]{IEEEtran}
\usepackage{amsmath,amsfonts}
\usepackage{algorithmic}
\usepackage{algorithm}
\usepackage{array}
\usepackage[caption=false,font=normalsize,labelfont=sf,textfont=sf]{subfig}
\usepackage{textcomp}
\usepackage{stfloats}
\usepackage{url}
\usepackage{verbatim}
\usepackage{graphicx}
\usepackage{cite}
\usepackage{amssymb}
\usepackage{ragged2e}

\usepackage{booktabs}       
\usepackage{comment}
\usepackage[hidelinks]{hyperref}
\usepackage{microtype}      
\usepackage{xcolor}         
\usepackage{soul}
\usepackage{balance}
\usepackage{float}
\usepackage{multirow} 
\usepackage{makecell}
\usepackage{enumitem}
\usepackage{color, colortbl}
\definecolor{Gray}{gray}{0.9}
\usepackage{caption}

\newtheorem{problem}{Problem Definition}

\begin{document}

\title{A Survey of Deep Graph Learning under Distribution Shifts: from Graph Out-of-Distribution Generalization to Adaptation}
\author{Kexin Zhang,
        Shuhan Liu,
        Song Wang, 
        Weili Shi,
        Chen Chen,
        Pan Li~\IEEEmembership{Member,~IEEE},
        Sheng Li~\IEEEmembership{Senior Member,~IEEE},
        Jundong Li~\IEEEmembership{Member,~IEEE},
        Kaize Ding*~\IEEEmembership{Member,~IEEE}

\IEEEcompsocitemizethanks{\IEEEcompsocthanksitem K. Zhang, S. Liu and K. Ding are with Department of Statistics and Data Science, Northwestern University, Evanston, IL, US.\protect\\
E-mail: kevin.kxzhang@gmail.com, shuhanliu@u.northwestern.edu, kaize.ding@northwestern.edu
\IEEEcompsocthanksitem S. Wang and J. Li are with Department of Electrical and Computer Engineering,
University of Virginia, Charlottesville, Virginia, US. J. Li is also with Department of Computer Science and School of Data Science, University of Virginia, Charlottesville, Virginia, US.\protect\\
E-mail: \{sw3wv, jundong\}@virginia.edu
\IEEEcompsocthanksitem W. Shi and S. Li are with School of Data Science, University of Virginia, Charlottesville, Virginia, US.\protect\\
E-mail: \{rhs2rr, shengli\}@virginia.edu
\IEEEcompsocthanksitem C. Chen is with Department of Computer Science, University of Virginia, Charlottesville, Virginia, US.\protect\\
E-mail: zrh6du@virginia.edu
\IEEEcompsocthanksitem P. Li is with School of Electrical and Computer Engineering, Georgia Institute of Technology, Atlanta, GA, US, and with Department of Computer Science, Purdue University, West Lafayette, IN, US.\protect\\
E-mail: panli@gatech.edu
}

\thanks{†Kexin Zhang is a research intern during the completion of this work.\protect\\
*Kaize Ding is the corresponding author.}
}

\markboth{Journal of \LaTeX\ Class Files,~Vol.~14, No.~8, August~2021}%
{Shell \MakeLowercase{\textit{et al.}}: A Sample Article Using IEEEtran.cls for IEEE Journals}

\maketitle

\begin{abstract}
Distribution shifts on graphs -- the discrepancies in data distribution between training and employing a graph machine learning model -- are ubiquitous and often unavoidable in real-world scenarios. These shifts may severely deteriorate model performance, posing significant challenges for reliable graph machine learning. Consequently, there has been a surge in research on graph machine learning under distribution shifts, aiming to train models to achieve satisfactory performance on out-of-distribution (OOD) test data. In our survey, we provide an up-to-date and forward-looking review of deep graph learning under distribution shifts. Specifically, we cover three primary scenarios: graph OOD generalization, training-time graph OOD adaptation, and test-time graph OOD adaptation.
We begin by formally formulating the problems and discussing various types of distribution shifts that can affect graph learning, such as covariate shifts and concept shifts. 
To provide a better understanding of the literature, 
we introduce a systematic taxonomy that classifies existing methods into model-centric and data-centric approaches, investigating the techniques used in each category.
We also summarize commonly used datasets in this research area to facilitate further investigation. Finally, we point out promising research directions and the corresponding challenges to encourage further study in this vital domain. 
We also provide a continuously updated reading list at \url{https://github.com/kaize0409/Awesome-Graph-OOD}.
\end{abstract}


\section{Introduction}\label{sec:introduction}
\IEEEPARstart{D}{riven} by the prevalence of graph-structured data in numerous real-world scenarios, growing attention has been paid to graph machine learning, which effectively captures the relationships and dependencies among entities within graphs.
In particular, Graph Neural Networks (GNNs) have emerged as powerful tools for learning representations on graphs through message-passing~\cite{Kipf2016Semi,wu2019simplifying,hamilton2017inductive}, and they have demonstrated remarkable success across diverse applications, such as social networks, physics problems, and traffic networks~\cite{Bi2023PredictingTS,Liu2023StructuralRI,Zhu2020TransferLO}.

While graph machine learning has achieved notable success, most of the existing efforts assume that test data follows the same distribution as training data, which is often invalid in the wild.
When confronted with Out-Of-Distribution (OOD) samples, the performance of graph machine learning methods may substantially degrade, limiting their efficacy in high-stake graph applications such as finance and healthcare \cite{Li2022OutOfDistributionGO}.
Although numerous transfer learning methods have been proposed to address distribution shifts for Euclidean data \cite{Zhuang2020ACS,liang2023comprehensive,fang2022source},
their direct application to graph data is challenging. This is due to the interconnected nature of entities on graphs, which violates the independent and identically distributed (IID) assumption inherent in traditional transfer learning methods.
Moreover, the various types of graph shifts introduce new challenges.
These shifts occur across different aspects including features, structures, and labels, and can manifest in various forms such as variations in graph sizes, subgraph densities, and homophily \cite{chen2022learning}.
Given these obstacles,
increasing research efforts have been dedicated to improving the reliability of graph machine learning against distribution shifts, concentrating on three main scenarios: graph OOD generalization ~\cite{Li2022OutOfDistributionGO,chen2022learning}, training-time graph OOD adaptation~\cite{Guo2023LearningAN, Qiao2023SemisupervisedDA}, and test-time graph OOD adaptation~\cite{Wang2022TestTimeTF, zhang2024collaborate}.

The primary distinction between graph OOD generalization and adaptation methods lies in their assumptions regarding the availability of target data.
Graph OOD generalization assumes the unavailability of target data during model training and aims to enhance the model's generalization performance on potentially unseen test distributions.
In contrast, both training-time and test-time adaptation assume the availability of target data and focus on improving model performance for a specific target domain.
However, they differ in their use of source data and how they leverage source distribution knowledge.
Training-time adaptation assumes that both the source and target graphs are available simultaneously, allowing the model adaptation to start from scratch during the training process.
On the other hand, test-time adaptation typically assumes access to a model pre-trained on the source domain, rather than the source graph itself, and begins adapting the model to the target data from this pre-trained state. Although graph OOD generalization, training-time OOD adaptation, and test-time OOD adaptation are closely related, there is currently no unified framework that comprehensively discusses deep graph learning under distribution shifts across all three scenarios.

With recent progress on graph OOD learning, an up-to-date and forward-looking review of this field is urgently needed. 
In this survey, we provide, to the best of our knowledge, the first unified and systematic review of the literature on deep graph learning under distribution shifts.
We start by formally formulating the problems and discussing different types of graph distribution shifts in graph machine learning.
Next, our new taxonomy is proposed, classifying existing methods into three categories based on the model learning scenario: 
(1) \textit{graph OOD generalization}, 
which typically relies on strong regularization techniques and domain-invariant features to generalize well to unseen domains,
(2) \textit{training-time graph OOD adaptation}, 
which offers a more flexible approach by allowing access to both source and target data, making it more suitable for situations where domain shift is expected but some target data is available \cite{You2023GraphDA,Zhu2023ExplainingAA}, and
(3) \textit{test-time graph OOD adaptation}, 
which does not have access to the source data during adaptation and focuses on leveraging a pre-trained source model to adjust to the target domain at test time. This approach is especially useful in online learning or real-time adaptation scenarios \cite{Jin2022EmpoweringGR,Zhu2023GraphControlAC}.

To deepen our understanding of these approaches, we further classify existing methods within each of the three scenarios into model-centric and data-centric strategies. Model-centric approaches focus on improving the learning process or designing the architecture of the graph model.
These methods enhance the model's ability to generalize or adapt to distribution shifts by refining key aspects such as network structure, training objectives, or learning mechanisms.
On the other hand, data-centric approaches emphasize the manipulation of input graphs, improving model performance by addressing the data directly, either through preprocessing techniques or data augmentation strategies.
Within each subfield of research, we systematically analyze the detailed techniques designed to enhance either the generalizability or adaptability of graph models under distribution shifts.
Furthermore, we provide a summary of the datasets utilized in these studies, highlighting their characteristics and relevance to the challenges posed by distribution shifts.
Despite the progress in graph OOD generalization and adaptation, several open challenges remain.
In the end we also point out several promising research directions in this evolving field.

\textbf{Differences between this survey and existing ones.} 
There is a growing need for a comprehensive review of graph learning under distribution shifts, while existing surveys have primarily focused on specific subdomains rather than providing a unified perspective across different scenarios.
Until now, 
several related surveys have been conducted, each with a distinct focus,
including graph OOD generalization~\cite{Li2022OutOfDistributionGO, ma2024survey}, graph domain adaptation~\cite{zhao2022graph, shi2024graph}, trustworthy graph learning~\cite{ju2024survey} related to distribution shifts. Our work differs from these surveys in the following key aspects:
(1) \textbf{Main focus}. 
We provide an in-depth analysis of both challenges and methodological advancements in graph learning under distribution shifts. In contrast, \cite{ju2024survey} analyzes OOD issues from a trustworthy perspective but does not delve into the methodological aspects. Similarly, \cite{ma2024survey} approaches the problem through a causal lens, which, while valuable, offers a narrower perspective compared to our broader investigation. 
(2) \textbf{Taxonomy}. We introduce a comprehensive categorization of existing methods and systematically summarize them.
In comparison, prior surveys such as \cite{wu2024graph} lack a comprehensive taxonomy, while \cite{zhao2022graph} and \cite{shi2024graph} primarily concentrate on domain adaptation, without addressing the broader landscape of graph OOD learning.
Additionally, we provide coverage of the most recent advancements and discussions in this field, capturing emerging trends, newly proposed techniques, and ongoing challenges. 

\textbf{Survey structure.} The general organization of this survey is presented as follows: Section~\ref{sec:Pre} introduces the notations and preliminaries. Sections \ref{sec:OOD}, \ref{sec:TrTA} and \ref{sec:TTA} review graph OOD generalization, training-time graph OOD adaptation, and test-time graph OOD adaptation, respectively. Each section discusses model-centric and data-centric approaches within its scenario, further detailing the techniques associated with each category. Section \ref{sec:dataset} provides a comprehensive summary of the datasets used in the literature, highlighting popular graph datasets for evaluation and their relevance to the challenges posed by distribution shifts. Section \ref{sec:future} explores promising future research directions and the associated challenges. Finally, Section \ref{sec:conclusion} presents the conclusion of this survey.

\section{Preliminaries}
\label{sec:Pre}
\subsection{Problem Definition}

Let $\mathcal{V}$ denote the node set with $N$ nodes in a graph $\mathcal{G}=(\textbf{A}, \textbf{X})$, where $\textbf{A} \in \mathbb{R}^{N \times N}$ is the adjacency matrix, with $a_{u v}$ representing the connection between nodes $i$ and $j$. The node feature matrix $\textbf{X} \in \mathbb{R}^{N \times F}$ contains $F$-dimensional feature vectors for each node $v \in \mathcal{V}$. Additionally, $\textbf{Y} \in \mathbb{R}^{N \times K}$ represents the node label matrix, where each label $y_v$ corresponds to one of the $K$ classes.

In this section, we take node classification as an exemplar to define node-level tasks of graph learning under distribution shifts. 
However, these definitions can be readily extended to other tasks, such as edge-level and graph-level learning tasks, which are also discussed throughout the survey.
Specifically, let $\varphi_{\theta}$ denote a graph model characterized by parameters $\theta$. 
Generally, the deep graph model $\varphi$ can be decomposed as $g \circ f$, where $g (\cdot): (\textbf{A}, \textbf{X}) \rightarrow \textbf{H}$ is a graph encoder mapping the graph inputs to latent representations $\textbf{H}$, and $f(\cdot)$ is a classifier in the latent space. We summarize the notations in Table~\ref{tab:notations}.
\begin{table}[ht]
\centering
\caption{Frequently used notations.}
\scalebox{0.9}{
\begin{tabular}{c|l}
\toprule
\textbf{Notation} & \textbf{Description} \\
\toprule
$\mathcal{G}=(\textbf{A}, \textbf{X})$ & The (original) graph. \\
$\mathcal{V}$ & The set of nodes. \\
$\textbf{A}$ & The adjacency matrix. \\
$\textbf{X}$ & Node feature matrix. \\
$\textbf{Y}$ & Node label matrix. \\
$\textbf{H}$ & The latent representations of the graph. \\
$K$ & The number of classes in the node classification task. \\
$N$ & The number of nodes. \\
$\mathcal{P}_{S}, \mathcal{P}_{T}$ & Source and target distributions. \\
$\varphi_{\theta}$ & The graph model characterized by parameters $\theta$. \\
$g(\cdot)$ & The graph encoder function. \\
$g_s(\cdot)$ & The graph encoder that extracts invariant task-related information. \\
$g_o(\cdot)$ & The graph encoder used for other components, excluding $g_s$. \\
$f(\cdot)$ & The classifier function. \\
$f_d(\cdot)$ & The domain discriminator function. \\
$l(\cdot)$ & The learning objective. \\
\bottomrule
\end{tabular}}
\label{tab:notations}
\end{table}

\begin{figure}[t!] 
\centering
\scalebox{0.99}{\includegraphics[width=0.48\textwidth]{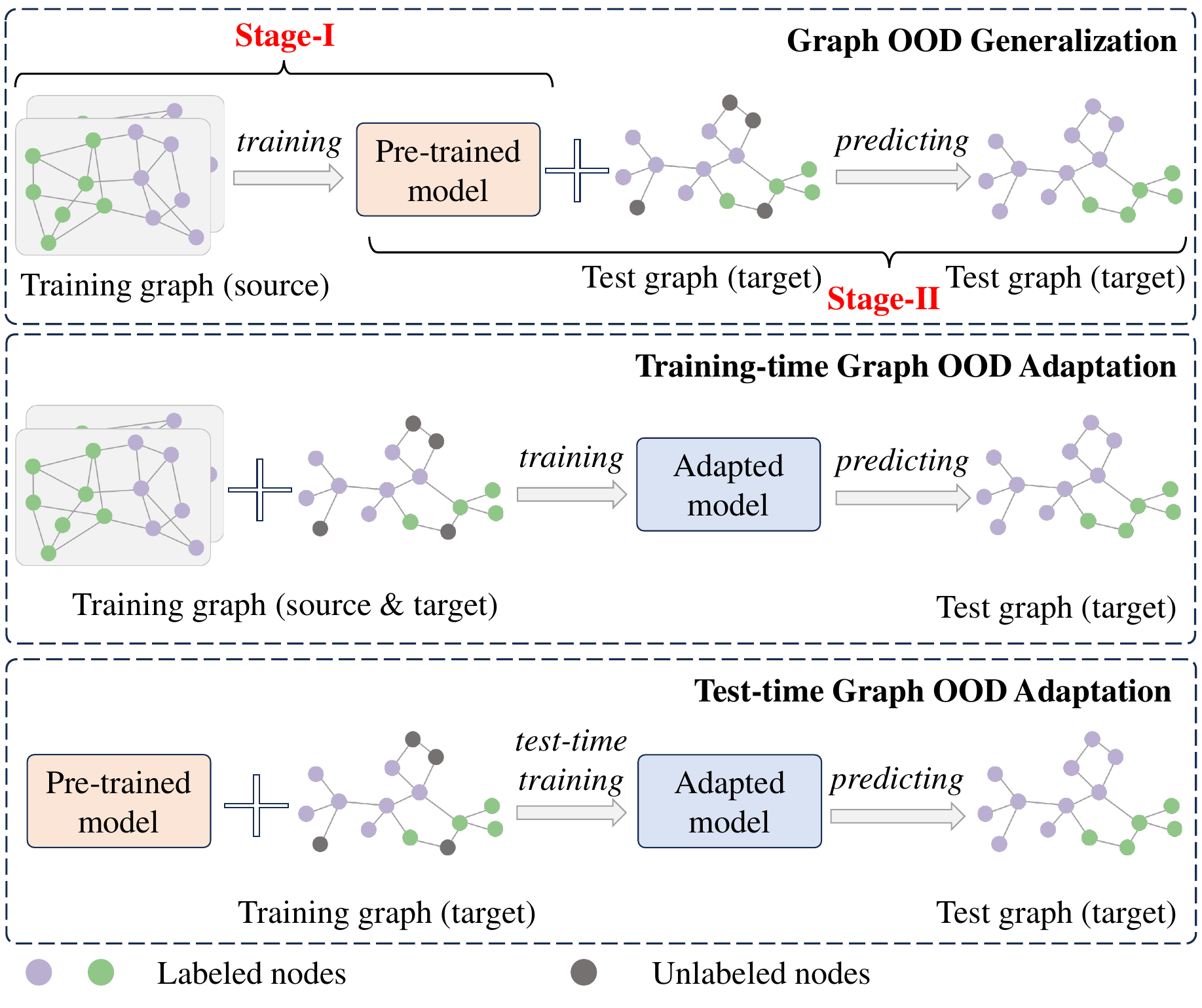}}
\caption{An illustration of the distinctions between graph OOD generalization, training-time graph OOD adaptation, and test-time graph OOD adaptation. The primary difference lies in the availability of target data and how source data is utilized. Graph OOD generalization assumes no target data during training, while both training-time and test-time adaptations assume access to target data. Training-time adaptation uses both source and target data during training, whereas test-time adaptation adapts a pre-trained model with only target data. The source graph can be single or multiple.}
\label{Fig:three_scenarios}
\end{figure}

We consider three distinct problem scenarios (i.e., \textit{Graph OOD Generalization}, \textit{Training-time Graph OOD Adaptation}, and \textit{Test-time Graph OOD Adaptation}) and the illustration can be found in Figure~\ref{Fig:three_scenarios}.

\begin{problem}
\textbf{Graph OOD Generalization}:
In this scenario, we assume that distribution shifts exist between source and target distributions, where $\mathcal{P}_{S} \neq \mathcal{P}_{T}$, and that $\mathcal{P}_{T}$ is not observable during model training.
The objective of node-level graph OOD generalization is to learn an optimal model $\varphi_{{\theta}^*}$ using the nodes and labels from the source graph $(\mathcal{G}_S, \textbf{Y}_{S})$,
such that it generalizes well to the unseen target graph $\mathcal{G_T}$.
\begin{equation}
\label{eq:problem}
    \varphi_{{\theta}^*} = \arg \min _{\theta} \mathbb{E}_{\mathcal{P}_{S}} [l(\varphi_{\theta}(\mathcal{G}_S), \textbf{Y}_{S})], 
\end{equation}
where $l(\cdot)$ is the loss function that measures the discrepancy between the model's predictions $\varphi_{\theta}(\mathcal{G}_S)$ and the ground-truth node labels $\textbf{Y}_S$. 
The learned model $\varphi_{{\theta}^*}$ is then used to infer node labels $\hat{\textbf{Y}}_{T} = \varphi_{{\theta}^*}(\mathcal{G}_T)$ for the target graph.

\end{problem}

\begin{problem}
\textbf{Training-time Graph OOD Adaptation}:
This scenario involves a source graph $\mathcal{G}_S$ 
with corresponding node labels $\textbf{Y}_S$, and a target graph $\mathcal{G}_T$ with node labels $\textbf{Y}_T$, where $\textbf{Y}_T$ may be partially labeled or entirely unlabeled (i.e., $\textbf{Y}_T = \varnothing$).
Under the assumption that there exist distribution shifts between the source and target distributions, the objective of node-level training-time graph OOD adaptation is to learn an optimal model $\varphi_{{\theta}^*}$ that minimizes the loss on the target graph by leveraging both the source and target graphs:
\begin{equation}
    \varphi_{{\theta}^*} = \arg \min _{\theta} \mathbb{E}_{\mathcal{P}_{S}, \mathcal{P}_{T}} [l(\varphi_{\theta}(\mathcal{G}_S, \mathcal{G}_T), \textbf{Y}_S, \textbf{Y}_T)]. 
\end{equation}

If $\textbf{Y}_T = \varnothing$, indicating that none of the nodes in the target graph are labeled, the problem is classified as unsupervised domain adaptation.
In this case, the learning objective $l(\cdot)$ relies solely on the labeled source domain data without direct supervision from the target domain.
Otherwise, when a portion of the nodes in the target graph are labeled, the problem is considered semi-supervised, and $l(\cdot)$ can utilize both the source domain labels and the partial target domain labels to optimize model performance.
\end{problem}

\begin{problem}
\textbf{Test-time Graph OOD Adaptation}:
In this scenario, a pre-trained model $\varphi_{\theta^\prime}$ and a target graph $\mathcal{G_T}$ with node labels $\textbf{Y}_T$ (which may be partially labeled or entirely unlabeled) are given. The goal is to adapt the pre-trained model during the test-time training phase in response to distribution shifts in the target domain. This process is performed without access to source domain data during adaptation. Given the distribution shifts between the source and target domains, the objective is to adjust the pre-trained model $\varphi_{\theta^\prime}$ at test time to optimize its performance on the target graph:
\begin{equation}
    \varphi_{{\theta}^*} = \arg \min _{\theta} \mathbb{E}_{\mathcal{P}_{T}} [l(\varphi_{\theta^{\prime}, \theta}(\mathcal{G}_T), \textbf{Y}_T)]. 
\end{equation}

Here, $\theta^{\prime}$ represents the parameters from the pre-trained model, while $\theta$ denotes the parameters introduced for test-time adaptation.
\end{problem}

\noindent
\subsection{Graph Distribution Shift Types}
In traditional machine learning, several studies have discussed and defined various types of distribution shifts \cite{moreno2012unifying, kull2014patterns}, of which the most widely-used concepts are \textit{covariate shifts} (i.e., shifts in $\mathcal{P}(\textbf{X})$) and \textit{concept shifts} (i.e., shifts in $\mathcal{P}(\textbf{X}|\textbf{Y})$ or $\mathcal{P}(\textbf{Y}|\textbf{X})$).
These concepts describe how data distributions change between training and test phases or across different domains.
These shifts can naturally be extended to graph setting by replacing feature inputs $\textbf{X}$ with the graph inputs $\mathcal{G}=(\textbf{A}, \textbf{X})$.

\textbf{Covariate Shifts.}
Covariate shifts on graphs emphasize the changes in graph inputs  $\mathcal{P}(\textbf{A}, \textbf{X})$. This can be further decomposed and interpreted as \textit{structure covariate shift}, \textit{size covariate shift}, and \textit{feature covariate shift}~\cite{Li2022OutOfDistributionGO}. Specifically, size covariate shift refers to changes in the graph size distribution, which can be captured as shifts in $\mathcal{P}(\textbf{A}, \textbf{X}, N)$, where $N$ denotes the graph size~\cite{yehudai2021local}.

\textbf{Concept Shifts.}
Concept shifts on graphs highlight the shifts in the relationship between graph inputs and labels $\mathcal{P}(\textbf{Y}|\textbf{A}, \textbf{X})$ or $\mathcal{P}(\textbf{A}, \textbf{X}|\textbf{Y})$. 
The concept shifts can be further decomposed to reveal more specific graph distribution shift types, such as feature concept shift $\mathcal{P}(\textbf{X}|\textbf{Y})$ and structure concept shift $\mathcal{P}(\textbf{A}|\textbf{Y})$ \cite{Liu2023StructuralRI}.

The usage of these concepts often extends beyond the input space to the latent representation space $\textbf{H}$, with  
covariate shifts describing the distribution shifts in latent representations $\mathcal{P}(\textbf{H})$, and concept shifts describing the changes in $\mathcal{P}(\textbf{H}|\textbf{Y})$ or $\mathcal{P}(\textbf{Y}|\textbf{H})$.

The impact of these shifts on graph learning tasks is multifaceted. Covariate shifts, particularly in graph structure, can lead to severe performance degradation, as models might not effectively learn the underlying relationships between nodes when their connectivity patterns differ. Concept shifts, on the other hand, often lead to misalignments in the model’s assumptions about the feature-label or structure-label dependencies. This issue is particularly problematic in tasks that heavily rely on the specific relationships between the graph components, such as link prediction or node classification.

\begin{figure*}[t!] 
\centering
\includegraphics[width=\textwidth]{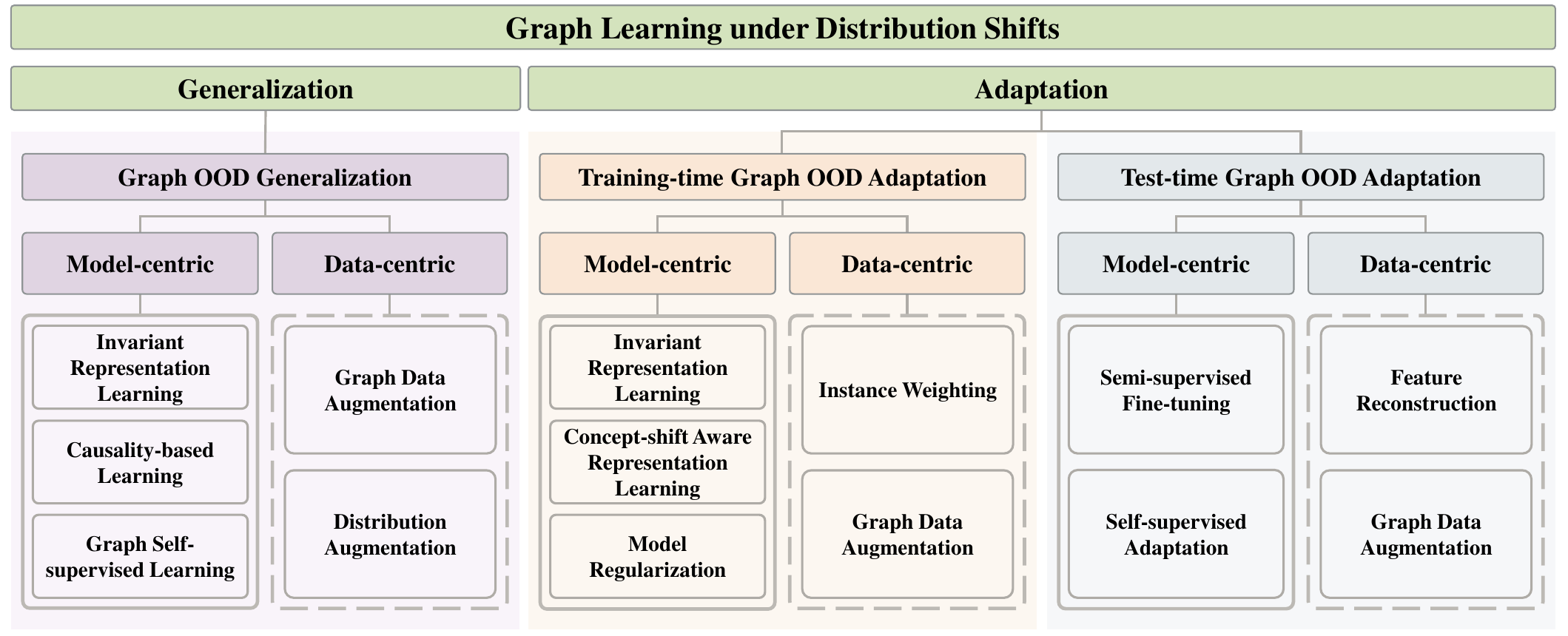}
\caption{Overview of our proposed taxonomy. Within each of the three scenarios, existing methods can be further categorized into model-centric and data-centric strategies.}
\label{Fig:tax}
\end{figure*}

\subsection{Discussion on Related Topics}
Several topics are closely related to graph OOD learning, including: graph transfer learning, graph domain adaptation, graph OOD detection, and graph continual learning. 
While these topics share similar goals with graph OOD learning, they differ in scope and methodology. Below, we discuss their distinctions and connections.

\noindent
\textbf{Graph Transfer Learning.} 
Traditionally, transfer learning is a more wide-ranging concept and involves transferring knowledge across distribution changes and diverse tasks. In contrast to the focus of this survey—graph learning under distribution shifts—graph transfer learning is not solely concerned with addressing distribution shifts. It also includes leveraging 
learned representations, model parameters, or structural patterns
gained from one task to enhance performance in a different task.

\smallskip
\noindent
\textbf{Graph Domain Adaptation.}
Following traditional domain adaptation, graph domain adaptation methods~\cite{zhao2022graph, shi2024graph} typically rely on the covariate shift assumption, which posits an invariant relationship between graph inputs and labels, represented as $\mathcal{P}_S(\textbf{Y}|\textbf{A}, \textbf{X})=\mathcal{P}_T(\textbf{Y}|\textbf{A}, \textbf{X})$.
The objective is to address distribution shifts in the input graph space across domains, i.e., $\mathcal{P}_S(\textbf{A}, \textbf{X}) \neq \mathcal{P}_T(\textbf{A}, \textbf{X})$. 
In contrast to graph domain adaptation, graph OOD learning is more general and comprehensive, involving distribution shifts that go beyond the covariate shift assumption.

\smallskip
\noindent
\textbf{Graph OOD Detection.} 
Graph OOD detection aims to empower models to recognize samples that significantly deviate from the distribution of the training data. 
In particular, the open-set problem, wherein unseen classes emerge in the test data, can be regarded as a special case of the OOD detection problem by defining the OOD nodes as nodes with labels unseen from the training set \cite{song2022learning}. 
Tackling distribution discrepancy in the open-set setting typically involves the dual objective of performing OOD detection to identify uncertain nodes while also adapting the model to the remaining non-OOD nodes \cite{wang2024open, wang2024goodat}.

\smallskip
\noindent
\textbf{Graph Continual Learning.} 
Graph continual learning involves adapting to a continuous stream of evolving graph data while retaining previously acquired knowledge~\cite{zhang2024continual,qi2024revealing}. This process focuses on the model's ability to incrementally learn and update itself as new data arrives, without forgetting past information. In contrast, graph OOD learning is primarily concerned with managing distribution shifts in static data and is focused on specific scenarios where such shifts occur.

\subsection{Proposed Taxonomy}
As highlighted in the problem definitions, graph OOD generalization, training-time graph OOD adaptation, and test-time graph OOD adaptation, significantly differ in their learning settings.
While these scenarios are 
related by their common focus on handling distribution shifts in graph learning,
they represent different stages of dealing with distribution shifts: \textit{from generalization with no access to target data, to training-time adaptation where both domains are considered, to test-time adaptation when the pre-trained model and target domain data are available.} Understanding these relationships helps form a more coherent view of how models can adapt to shifting data distributions.
Accordingly, in the following three sections, we firstly categorize existing methods into graph OOD generalization, training-time graph OOD adaptation, and test-time graph OOD adaptation.
Within each section, we further classify methods into model-centric and data-centric approaches.
Model-centric approaches center on the learning process or the design of the graph model, while data-centric approaches emphasize the manipulation of input graphs, such as transforming graph structures or features.

Our taxonomy is illstrated in Figure \ref{Fig:tax}. Note that even though the method categories for the three scenarios—graph OOD generalization, training-time graph OOD adaptation, and test-time graph OOD adaptation—may converge in terms of their high-level strategies (e.g., invariant representation learning, graph data augmentation), their application and emphasis on specific aspects, such as the data availability and the stage of the learning process, are what distinguish them.

\section{Graph OOD Generalization}
\label{sec:OOD}
Graph OOD generalization addresses the challenge of learning models that remain effective under distribution shifts, where the test data follows a different distribution from the training data~\cite{fan2022debiasing, Gui2023Joint, Li2023GraphSA, guo2024investigating}.
The setting of graph OOD generalization assumes that the model has no access to the target domain during training, making generalization particularly challenging~\cite{Liu2023StructuralRI, Jin2022EmpoweringGR}. 
The goal is to develop models that can capture domain-invariant patterns and generalize across diverse, unseen target distributions without 
prior knowledge of the target domain's characteristics~\cite{Yu2023MindTL}.
In this section, we provide a comprehensive review of the most recent advances in graph OOD generalization, highlighting both model-centric and data-centric approaches. A summary of existing methods is presented in Table~\ref{tab:ood}.

\subsection{Model-Centric Approaches}
In this section, we introduce model-centric approaches for graph OOD generalization, which focus on the strategies in model learning to improve generalizability. Based on specific designs and objectives, these approaches could be further categorized into \textit{invariant representation learning},  \textit{causality-based learning} and \textit{graph self-supervised learning}.

\smallskip 
\noindent\textbf{Invariant Representation Learning.} Invariant representation learning methods for graph OOD {generalization} 
aim to extract invariant information that is robust to distribution shifts across different environments~\cite{zhu2023mario,sadeghi2021distributionally, feng2019graph}. Here, "environments" refer to distinct settings that can influence the structure and relationships within the graph data. These methods can be further classified into \textit{(1) invariant subgraph extraction}, \textit{(2) adversarial learning}, and \textit{(3) disentangled representation learning}. 
These subcategories reflect specific learning techniques to learn invariant features from graph data.
To provide a clearer understanding of these methods, we illustrate their key ideas in Figure~\ref{Fig:invariant}.

\begin{itemize}[wide]
\item \textit{(1) Invariant Subgraph Extraction.} 
Inspired by the invariant learning strategy in traditional OOD generalization tasks~\cite{muandet2013domain,arjovsky2019invariant,ahuja2020invariant,creager2021environment}, invariant subgraph extraction methods divide each graph into two distinct segments~\cite{li2022learning,miao2022interpretable}: an invariant subgraph, which maintains stable structure-label relationships across different distributions, and a variant subgraph, which may change across distributions. 
By concentrating on the invariant subgraph, these methods 
enable models to learn representations that levearge stable graph-label dependencies rather than distribution-specific variations, thereby improving its generalization performance. For example,
DIR~\cite{wu2022discovering} exemplifies this approach by distinguishing invariant subgraphs using a GNN-based approach, enabling the identification of consistent graph-label relations without the need for explicit distribution-based separation in the training data~\cite{wu2022discovering}. 
Conversely, GIL~\cite{li2022learning} advances this concept by utilizing the distributional divisions of training data to cluster variant subgraphs, thereby facilitating the generation of invariant graph representations for accurate classification. 
Additionally, GSAT~\cite{miao2022interpretable} formulates the graph Information Bottleneck (IB) principle~\cite{tishby2015deep,wu2020graph} by injecting learnable randomness into a node-level attention mechanism, which allows selecting subgraphs that keep invariant to the environmental shift and indicative to the labels. 

A recent work by Piao et al.~\cite{piao2024improving} proposes a hierarchical approach for effective subgraph extraction. This method constructs multiple semantic environments for each graph while ensuring consistency across hierarchical levels. 
By considering multiple environments and modeling the correlations between different environments within a hierarchical framework, this approach can better understand the inherent variability in the data, leading to improved robustness against distribution shifts.
Sun et al.~\cite{sun2024dive} propose DIVE framework, which trains multiple models on all label-predictive subgraphs, consisting of both spurious and invariant subgraphs. This encourages the models to focus on distinct structural patterns within these subgraphs, facilitating the precise extraction of invariant subgraphs and mitigating the effects of simplicity bias.

\begin{figure}[t!] 
\centering
\scalebox{0.99}{\includegraphics[width=0.48\textwidth]{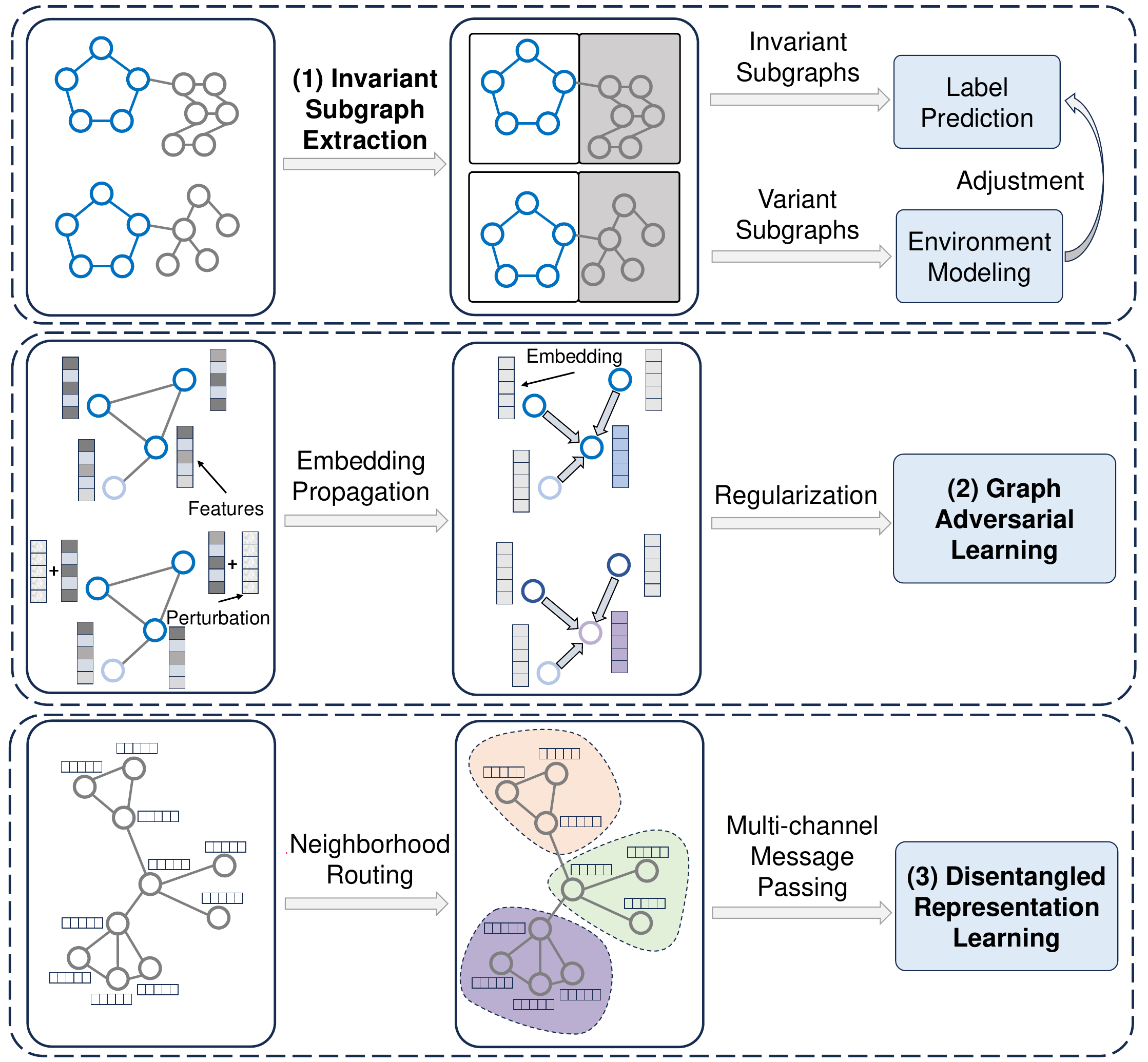}}
\caption{Illustration of three invariant representation learning approaches: (1) \textit{Invariant Subgraph Extraction} partitions graph structures for stability, (2) \textit{Adversarial Learning} enhances robustness via perturbations, and (3) \textit{Disentangled Representation Learning} separates independent factors for better generalization.}
\label{Fig:invariant}
\end{figure}

\item \textit{(2) Adversarial Learning.}
Adversarial learning methods aim to learn invariant representations in different environments by enforcing models to perform consistently 
under adversarially induced distribution shifts~\cite{wu2019domain,sadeghi2021distributionally}. 
These methods leverage adversarial perturbations or optimization strategies to strengthen the model’s ability to handle unseen distribution changes.
Pioneering works in adversarial training have adapted existing adversarial techniques from image-based OOD generalization to graph learning.
For example, GNN-DRO~\cite{sadeghi2021distributionally} employs distributionally robust optimization~\cite{rahimian2019distributionally} to address OOD challenges, aiming to minimize the worst-case loss within a defined range around the observed data distribution.
However, the reliance of GNN-DRO on distribution labels can limit its application in the real world~\cite{wu2022handling}. To address this limitation, GraphAT~\cite{feng2019graph} introduces graph perturbations to create augmented distributions. These perturbations are achieved by maximizing the prediction discrepancies between a given node and its neighboring nodes, thereby effectively altering the graph's smoothness to generate more diverse distributions. 

\begin{table*}[ht]
\setlength\tabcolsep{25.pt}
\centering
\caption{A summary of graph OOD generalization methods. `Task Level' denotes the main task level, `Domain Source' denotes whether the source domain consists of single or multiple domains, and `Distribution Shift' denotes the distribution shifts that the methods aim to handle.}
\scalebox{0.6852}{
\begin{tabular}{c|c|ccccc}
\bottomrule
\rowcolor{Gray}  &  &  &  &  &  &  \\
\rowcolor{Gray}  
\multirow{-2}{*}{\textbf{Scenario}} &\multirow{-2}{*}{\textbf{Approach}} & \multirow{-2}{*}{\textbf{Technique}} & \multirow{-2}{*}{\textbf{Method}} & \multirow{-2}{*}{\makecell{\textbf{Task} \\ \textbf{Level}}} & \multirow{-2}{*}{\makecell{\textbf{Domain} \\ \textbf{Source}}} & \multirow{-2}{*}{\makecell{\textbf{Distribution} \\ \textbf{Shift}}}  \\ \toprule
\multirow{64}{*}{\rotatebox{90}{\makecell{Graph OOD Generalization}}} &
\multirow{45}{*}{\rotatebox{90}{\makecell{Model-centric}}} 
& \multirow{8}{*}{\makecell{Invariant\\Structure Extraction}}
& DIR~\cite{wu2022discovering} & Graph &Multiple&Covariate  \\\cmidrule{4-7}
& & & GIL~\cite{li2022learning} & Graph &Multiple&Covariate   \\\cmidrule{4-7}
& & & GSAT~\cite{miao2022interpretable} & Graph &Multiple&Covariate   \\\cmidrule{4-7}
& & & IS-GIB~\cite{yang2023individual} & Node/Graph&Single&Covariate  \\\cmidrule{4-7}
& & &DIVE~\cite{sun2024dive}& Graph& Multiple&Covariate \\\cmidrule{3-7}
& & \multirow{5}{*}{\makecell{Adversarial Learning}} 
& CAP~\cite{xue2021cap}& Node&Single &Covariate  \\\cmidrule{4-7}
& & &GraphAT~\cite{feng2019graph}& Node& Single &Covariate  \\\cmidrule{4-7}
& & &GNN-DRO~\cite{sadeghi2021distributionally}& Node& Single&Covariate\&Concept   \\\cmidrule{4-7}
& & &WT-AWP~\cite{wu2023adversarial}& Node/Graph&Single&Covariate  \\\cmidrule{3-7}
& & \multirow{20}{*}{\makecell{Disentangled\\Representation Learning}} 
& DisenGCN~\cite{ma2019disentangled}& Node&Single &Covariate  \\\cmidrule{4-7}
& & &IPGDN~\cite{liu2020independence}& Node& Single &Covariate  \\\cmidrule{4-7}
& & &FactorGCN~\cite{yang2020factorizable}& Node/Graph& Single&Covariate   \\\cmidrule{4-7}
& & &NED-VAE~\cite{guo2020interpretable}& Node/Edge& Single&Covariate   \\\cmidrule{4-7}
& & &DGCL~\cite{li2021disentangled}& Graph& Single&Covariate   \\\cmidrule{4-7}
& & &IDGCL~\cite{li2022disentangled}& Graph& Single&Covariate\&Concept   \\\cmidrule{4-7}
& & &I-DIDA~\cite{zhang2023out}& Node/Edge & Single&Covariate\&Concept   \\\cmidrule{4-7} 
& & &L2R-GNN~\cite{chen2024learning}& Graph & Single&Covariate \\\cmidrule{4-7} 
& & &OOD-GNN~\cite{li2022ood}& Graph & Single&Covariate   \\\cmidrule{4-7} 
& & & EQuAD~\cite{yao2024empowering} & Graph &Multiple&Covariate  \\\cmidrule{4-7}
& & &DIDA~\cite{zhang2022dynamic}& Node & Single&Covariate   \\\cmidrule{4-7} 
& & &EAGLE~\cite{yuan2024environment}& Node & Single&Covariate \\\cmidrule{4-7} 
& & &CauSTG~\cite{zhou2023maintaining}& Graph & Multiple&Covariate \\\cmidrule{3-7} 
& & \multirow{8}{*}{\makecell{Causality-based\\Learning}}
&E-invariant GR~\cite{bevilacqua2021size}&Graph&Single& Covariate (Size)  \\\cmidrule{4-7}
& & &CIGA~\cite{chen2022learning}&Graph&Single&Covariate\&Concept  \\\cmidrule{4-7}
& & &CAL~\cite{sui2022causal}&Graph&Single&Covariate  \\\cmidrule{4-7}
& & &DisC~\cite{fan2022debiasing} & Graph&Single&Covariate\&Concept  \\\cmidrule{4-7}
& & &GALA~\cite{chen2023does}& Graph&Single&Covariate\&Concept  \\\cmidrule{4-7}
& & &LECI~\cite{Gui2023Joint}& Graph&Single&Covariate  
 \\\cmidrule{4-7}
& & & StableGNN~\cite{fan2023generalizing}& Graph&Single&Covariate  
 \\\cmidrule{3-7}

& & \multirow{8}{*}{\makecell{Graph Self-supervised\\Learning}} 
& Pretraining-GNN~\cite{hu2020strategies} & Node/Graph &Single &Covariate  \\\cmidrule{4-7}
& & &PATTERN~\cite{yehudai2021local}& Node& Single &Covariate  \\\cmidrule{4-7}
& & &OOD-GCL~\cite{li2024disentangled}& Node/Graph& Single&Covariate   \\\cmidrule{4-7}
& & &GPPT~\cite{sun2022gppt}& Node/Graph& Single&Covariate\&Concept   \\\cmidrule{4-7}
& & &GPF~\cite{fang2024universal}& Node/Graph& Single&Covariate\&Concept   \\\cmidrule{4-7}
& & &GraphControl~\cite{zhu2024graphcontrol}& Node/Graph& Single&Covariate 
\\\cmidrule{2-7}
& \multirow{24}{*}{\rotatebox{90}{\makecell{Data-centric}}}
& \multirow{8}{*}{\makecell{Graph Perturbation}}
&AIA~\cite{Sui2023Unleashing}&Graph&Single&Covariate  \\\cmidrule{4-7}
& & &G-Splice~\cite{Li2023GraphSA}&Node/Graph&Single&Covariate\&Concept  \\\cmidrule{4-7}
& && LiSA~\cite{Yu2023MindTL}&Node/Graph&Single&Covariate\&Concept  \\\cmidrule{4-7}
& && DLG~\cite{wang2024enhancing}&Graph&Single&Covariate\&Concept  \\\cmidrule{4-7}
& && Pattern-PT~\cite{yehudai2021local}& Graph&Single & Covariate (Size)  \\\cmidrule{4-7}
& && P-gMPNN~\cite{zhou2022ood}& Edge&Single & Covariate (Size)  
\\\cmidrule{3-7}
& & \multirow{6}{*}{\makecell{Graph Mix-up}}
&GraphMix~\cite{verma2021graphmix}&Node&Single&Covariate  \\\cmidrule{4-7}
& && G-Mixup~\cite{wang2021mixup}&Node/Graph&Single&Covariate \\\cmidrule{4-7}
& && $\mathcal{G}$-Mixup~\cite{han2022g}&Graph&Single&Covariate \\\cmidrule{4-7}
& && OOD-GMixup~\cite{lu2023graph}&Node/Graph&Multiple&Covariate \\\cmidrule{3-7}
& & \multirow{10}{*}{\makecell{Distribution Augmentation}}
&GREA\cite{liu2022graph}&Graph&Single&Covariate \\\cmidrule{4-7}
& &&  EERM~\cite{wu2022handling} & Node &Multiple&Covariate \\\cmidrule{4-7}
& &&  FLOOD~\cite{liu2023flood} & Node &Multiple&Covariate \\\cmidrule{4-7}
& & &DPS~\cite{yu2022finding}&Graph&Single&Covariate \\\cmidrule{4-7}
& & &MoleOOD~\cite{yang2022learning}&Graph&Single&Covariate \\ \cmidrule{4-7}
& & &ERASE~\cite{chen2023erase}&Node&Single&Covariate \\ \cmidrule{4-7} 
& && IGM~\cite{jia2024graph}&Graph&Multiple&Covariate \\
\bottomrule
\end{tabular}
}
\label{tab:ood}
\end{table*}

\item \textit{(3) Disentangled Representation Learning.}
Disentangled representation learning aims to separate the distinct factors of variation within the data, where each factor captures a specific aspect of the graph structure. By isolating these independent factors, the approach ensures that they remain stable across different distribution shifts. This disentanglement allows the model to focus on the most invariant and task-relevant features of the data, thereby improving generalization across various domains or environments.
As a classic example, DisenGCN~\cite{ma2019disentangled} introduces a multichannel convolutional layer along with a neighborhood routing strategy to achieve node representation disentanglement, and the disentangled representations are more robust to distribution shifts regarding node labels. IPGDN~\cite{liu2020independence} enhances this disentanglement by promoting independence among latent factors based on Hilbert-Schmidt Independence Criterion (HSIC)~\cite{gretton2007kernel}, further improving generalization. From the perspective of factorization, FactorGCN~\cite{yang2020factorizable} proposes to produce disentangled representations on the graph level by breaking down graphs into multiple interpretable factor graphs. As such, the graph-level generalization is strengthened. 
In addition to the graph-level generalization, NED-VAE~\cite{guo2020interpretable} captures independent latent factors from both the node level and edge level with a variational autoencoder~\cite{kingma2019introduction,kipf2016variational}. This design improves generalization from various granularities of graph structures. 
Additionally, some works~\cite{chen2024learning,li2022ood,fan2023generalizing} point out that the degradation in OOD generalization ability  often stems from spurious correlations between irrelevant patterns and category labels within the training data. To mitigate this, L2R-GNN~\cite{chen2024learning} and OOD-GNN~\cite{li2022ood} utilize nonlinear graph decorrelation methods to eliminate spurious correlations between invariant patterns and other variables. L2R-GNN focuses on removing spurious correlations through a nonlinear approach, while OOD-GNN implements a global weight estimator for this purpose. Furthermore, EQuAD~\cite{yao2024empowering} proposes a graph invariance learning paradigm that disentangles invariant patterns from spurious features using the infomax principle~\cite{hjelmlearning,velivckovicdeep}, thus enhancing the interpretability and robustness of learned representations.

Another line of work has focused on utilizing contrastive learning
to achieve disentanglement. For example, DGCL~\cite{li2021disentangled} employs a self-supervised strategy to learn disentangled representations through factor-wise contrastive learning, circumventing the high computational cost associated with self-supervised graph reconstruction. 
IDGCL~\cite{li2022disentangled} improves DGCL by introducing independence regularization to remove dependencies among disentangled representations. Different from other methods, IDGCL learns multiple disentangled representations, which are more interpretable and robust to distribution shifts.

Beyond OOD generalization in conventional static graphs, dynamic graphs also face challenges from distribution shifts. Effectively modeling the complex latent environments and uncovering invariant patterns in dynamic graphs with distribution shifts remains an under-explored area. Zhang et al. propose DIDA~\cite{zhang2022dynamic}, a framework that utilizes a disentangled spatio-temporal attention network to differentiate between variant and invariant patterns, thereby improving predictive stability in dynamic graphs under distribution shifts. I-DIDA~\cite{zhang2023out} further refines predictive stability by identifying invariant patterns and applying disentangled spatio-temporal interventions.
Similarly, Yuan et al. develop the EAGLE framework~\cite{yuan2024environment}, which incorporates an environment-aware EA-DGNN to learn disentangled representations across various spatio-temporal environments. Zhou et al.~\cite{zhou2023maintaining} identify two key challenges in OOD generalization on spatiotemporal data: the entanglement of segment-level heterogeneity in temporal environments, and the failure of Invariant Risk Minimization~(IRM)~\cite{arjovsky2019invariant} on heterogeneous spatiotemporal observations. Their proposed CauSTG framework extracts invariant patterns through a spatiotemporal consistency learner and a hierarchical invariance explorer, directly addressing these challenges.

\end{itemize}

\noindent\textbf{Causality-based Learning.}
Rather than focusing on learning invariant or disentangled representations,
causality-based learning methods focus more on uncovering the inherent causal information in graph data crucial for classification and mitigating spurious correlations that could potentially mislead classification.
For instance, CIGA~\cite{chen2022learning} explores the causal information within each graph by extracting essential causal components related to labels, thereby enhancing the generalizability of the model.
DisC~\cite{fan2022debiasing} adopts the GCE loss~\cite{lee2021learning} to learn spurious information with a larger 
step size in order to capture the remaining causal information. 
To address the limitations of CIGA and DisC in scenarios where prior knowledge about the variance of causal and spurious information is unavailable, GALA~\cite{chen2023does} leverages an assistant model to detect distribution shifts in terms of such variance. It maximally learns the causal information based on proxy predictions, which helps in achieving successful OOD generalization under minimal assumptions. 
StableGNN~\cite{fan2023generalizing} tackles spurious correlations in subgraph units through causal analysis. It incorporates a differentiable graph pooling layer to obtain subgraph-based representations and applies a causal variable distinguishing regularizer to eliminate spurious correlations.
Recently, LECI~\cite{Gui2023Joint} proposes a two-stage approach for environment-based methods, consisting of two stages: environmental inference that focuses on inferring environmental labels, and environmental exploitation that focuses on leveraging these labels for effective data utilization. 
In particular, LECI integrates causal independence principles into a subgraph discovery framework. 
Apart from the general OOD generalization problem caused by domain shifts, 
E-invariant GR~\cite{bevilacqua2021size} 
specifically investigates the issue of size shifts. It proposes a causal model that learns representations that are more robust against variations in graph sizes.

\smallskip
\noindent\textbf{Graph Self-supervised Learning.} 
Graph self-supervised learning (SSL) has emerged as an effective strategy for improving graph OOD generalization by enabling models to learn transferable representations without relying on labeled data. The key objective is to enforce consistency in learned representations across different views or perturbations of the graph, ensuring that similar graphs are mapped to close representations~\cite{You2020GraphCL}. This enhances the model’s ability to generalize to unseen test graphs by capturing structural and semantic patterns that remain stable across distribution shifts~\cite{Li2022OutOfDistributionGO}.
Pretraining-GNN~\cite{hu2020strategies} is among the first to explore various graph self-supervised learning methods at both the node and graph levels to pre-train GNN models, including context prediction, attribute masking, and structural similarity prediction.  
To address graph size shifts, PATTERN~\cite{yehudai2021local} introduces a self-supervised task focused on learning d-pattern representations from local structures, thereby enhancing generalization from smaller to larger graphs.
Moreover, OOD-GCL~\cite{li2024disentangled} resorts to graph contrastive learning on factorized graph representations, thereby excluding the information from the latent factors sensitive to distribution shifts and improving overall generalization.

More recently, various Graph Foundation Models (GFMs)\cite{liu2023towards} have been developed based on graph self-supervised learning, demonstrating strong generalizability across diverse downstream tasks. One of the pioneering works in this area, GPPT\cite{sun2022gppt}, extends GNN models with prompt-tuning. Specifically, GPPT builds upon pre-trained GNNs trained via self-supervised learning and introduces additional tokens representing candidate labels, enabling better generalization to new tasks without requiring extensive fine-tuning.
Based on the prompt-tuning strategy, GPF~\cite{fang2024universal} proposes a universal prompt-tuning framework to pre-train GNN models across various self-supervised learning strategies. 
By operating in the feature space of input graphs, 
GPF improves the generalizability of the prompted graph to various downstream tasks in an adaptive way. Drawing inspiration from ControlNet~\cite{zhang2023adding}, GraphControl~\cite{zhu2024graphcontrol} enhances generalization with a universal structural pre-training strategy that enables the alignment of the input space across distributions and incorporates characteristics of target distributions as conditional contexts.

\subsection{Data-Centric Approaches}
\smallskip
\noindent

It is widely accepted that the generalization ability of a model heavily relies on the diversity and quantity of the training data \cite{Wang2021Gen, yue2024cooperative}. 
Therefore, one intuitive yet efficient way to boost the model's generalizability is to increase the diversity of training data through augmentation, including \textit{graph data augmentation} and \textit{distribution augmentation}.

\noindent\textbf{Graph Data Augmentation.}
Graph data augmentation directly modifies the structures and features of the graphs within the training set. This increases the diversity of the data, exposing the model to various graph topologies, node features, and edge relationships. By simulating variations of the input graphs, augmentation techniques allow the model to learn more robust and generalizable representations.
Based on the augmentation strategy, graph augmentation methods can be further classified into \textit{(1) graph perturbation} and \textit{(2) graph mix-up}.

\begin{itemize}[wide]
\item \textit{(1) Graph Perturbation.}
Graph perturbation methods involve small, controlled modifications to graph structures, such as adding or removing edges, or slightly altering node features, which introduces variability into the training data while preserving the original graph's core characteristics.
The augmented graphs, although slightly different from the originals, provide new and diverse training examples that capture the variability that the model might encounter in real-world applications.
By increasing the diversity of the graphs in the training set, graph perturbation methods help the model become more robust to a range of graph data distributions.
Specifically, the goal of graph data augmentation methods for OOD generalization could be formulated as follows:
\begin{equation}
\label{eq:augmentation}
\min_{f,g} \mathbb{E}_{\textbf{A}', \textbf{X}', \textbf{Y}'} [l(f ( g (\textbf{A}', \textbf{X}')), \textbf{Y}') ],
\end{equation}
where $\textbf{A}'$, $\textbf{X}'$, and $\textbf{Y}'$ represent the adjacency matrix, node feature matrix, and label of a perturbed graph, respectively. In this formulation, $g$ transforms the perturbed graph inputs into latent representations, while $f$ serves as the classifier.

As a classic example, Pattern-PT~\cite{yehudai2021local} studies the problem of graph size generalization and discovers discrepancies in small local structures across graphs can negatively impact the generalizability of GNNs. To mitigate this issue, Pattern-PT proposes an interpolation strategy among graphs in order to learn comprehensive representations of local structures. 
P-gMPNN~\cite{zhou2022ood} focuses on the link-level OOD generalization problem under graph size shifts.
This work explores the vulnerability of GNNs as link predictors based on structural node embeddings, and proposes a theoretically sound method that learns robust pairwise embeddings by modifying graphs into graphon fractions of common neighbors.
To prevent excessive modifications that might lead to information loss, several works aim to enhance the stability of augmentation strategies.
For instance, AIA~\cite{Sui2023Unleashing} considers the causal invariance principle while enlarging the diversity of environments, aiming to reduce the distribution gap between the augmented environments and the potential test data unseen during training. 
To ensure causal invariance, AIA designs a causal generator and adversarially learns the masks that encapsulate causal information. Similarly, G-Splice~\cite{Li2023GraphSA} is proposed to avoid the corruption of the underlying causal information related to labels. The authors explore the extrapolation strategies from both the perspectives of structures and features to augment graph data without impacting the causal information.
From a label-shift perspective, LiSA~\cite{Yu2023MindTL} 
identifies that augmented graph labels may change during structural modifications, 
and hence proposes to avoid such a label shift issue during graph augmentation.
Particularly, LiSA employs a variational subgraph generator, trained to produce label-invariant subgraphs, which are subsequently used in aiding graph OOD generalization.  DLG~\cite{wang2024enhancing} improves the label and distribution consistency by maximizing the mutual information between augmented and original graphs.

\item \textit{(2) Graph Mix-up.} 
The mix-up strategies, initially proposed and shown to be effective in computer vision and natural language processing, have demonstrated significant advantages in enhancing generalization through interpolating instances~\cite{zhang2018mixup,guo2020nonlinear,zhang2021does}. 
These techniques generate synthetic training graph data through mix-up, which enhances model robustness to distribution shifts and shows promise in graph OOD generalization~\cite{wu2021graphmixup}. For instance, GraphMix~\cite{verma2021graphmix} integrates manifold mixup~\cite{verma2019manifold} for generalizable node classification by simultaneously training a fully-connected network (FCN) and a GNN, with a parameter-sharing strategy that facilitates the transfer of essential node representations from the FCN to the GNN. Similarly, G-Mixup~\cite{wang2021mixup} extends this approach by interpolating latent node representations in the embedding space, offering generalizability across different graphs. 
Since latent representations may not fully capture the richness of the original data,
NodeAug~\cite{wang2020nodeaug} and ifMixup~\cite{guo2021ifmixup} further propose to directly interpolate raw node features and graph data, respectively, to create more diverse inputs for GNNs. 

On the other hand, $\mathcal{G}$-Mixup~\cite{han2022g} proposes to perform mixup on graphons across classes in the Euclidean space, resulting in new graphs derived from the interpolation of graphons.
The results demonstrate enhanced performance in graph classification tasks with distribution shifts and highlight the efficacy in promoting OOD generalization.
Instead of relying on predefined disturbances, OOD-GMixup~\cite{lu2023graph} generates virtual samples by applying manifold mixup directly to the graph rationale representations. This approach enables the simulation of complex graph distribution shifts by manipulating these representations within a metric space. Furthermore, to accurately measure and control the training process amid these shifts, OOD-GMixup introduces an OOD confidence score, derived using Extreme Value Theory (EVT). This score, coupled with a sample re-weighting mechanism, actively manages the training dynamics by prioritizing virtual OOD samples, thereby enhancing the model's generalization capabilities across varied and previously unseen graph distributions.
While extensive research efforts~\cite{han2022g,verma2021graphmix,wang2021mixup} have focused on interpolation for synthetic data augmentation, relatively few studies have explored the extrapolation in addressing out-of-distribution challenges on graphs. Recently, a novel framework~\cite{ligraph} was proposed to generate synthetic graphs using a non-Euclidean-space linear extrapolation method. The generated graphs have been theoretically validated to adhere to common causal assumptions. The framework consists of two components, which are designed to generate synthetic graphs in both the structural and feature spaces. Extensive experiments demonstrate that this extrapolation approach enhances a model's OOD generalization ability on graphs and outperforms previous invariant learning methods~\cite{li2022learning,wu2022discovering,chen2022learning} as well as existing graph data augmentation techniques~\cite{han2022g,verma2021graphmix,wang2021mixup}.
\end{itemize}

\noindent\textbf{Distribution Augmentation.} 
In contrast to graph data augmentation, which enhances the original graph at the graph level, distribution augmentation generates new distributions of graphs to facilitate robust representation learning~\cite{wu2022handling}.
This approach is particularly beneficial when the available training distribution is limited, making it difficult for the model to learn robust representations. In such cases, the model may struggle to distinguish between decisive information and spurious information that vary across distributions.

One of the pioneering methods in this domain is GREA~\cite{liu2022graph}, which is 
designed to address the challenge of scarce and limited graph distributions.
GREA operates under the assumption that each input graph contains a specific subgraph, known as the "graph rationale," that is crucial for making accurate predictions.
To leverage this, GREA augments graphs by replacing variant subgraphs (i.e., those not part of the graph rationale) in a target graph with variant subgraphs from other graphs. 
In this way, by training on these augmented graphs with varying domain subgraphs, the model will learn to focus solely on the graph rationale for making inferences, thereby enhancing OOD generalization.
Similarly, DPS~\cite{yu2022finding} generates diverse augmented subgraphs while preserving those containing the graph rationale. 
Multiple subgraph generators are used to produce additional distributions that capture different variations of the graph. To further improve the diversity of these augmented distributions, DPS incorporates a regularization term that increases the variance between them,
improving generalizability and robustness in the face of distributional changes.
Following this, IGM~\cite{jia2024graph} identifies invariant subgraphs and generates diverse environments by concatenating these with environment-related subgraphs. 
This co-mixup strategy not only creates diverse environments by mixing variant subgraphs, but also refines learning by mixing invariant subgraphs to eliminate spurious correlations and enhance OOD generalization.

Additionally, EERM~\cite{wu2022handling} introduces a distribution augmentation approach based on invariance principles. 
This approach leverages multiple context explorers, specifically graph structure editors, which are adversarially trained to maximize the variance in risk across the generated graph distributions. This design is particularly effective in dealing with the inherent complexity of graph structures, where node interconnections complicate direct augmentation.
Building on EERM, FLOOD~\cite{liu2023flood} further incorporates a bootstrapped learning module that could be refined on the test data to improve OOD generalization.
ERASE~\cite{chen2023erase} enhances generalization by employing a decoupled label propagation approach that corrects noisy labels through structural denoising before training. This method ensures that the model learns from more accurate representations of the data, effectively distinguishing meaningful information from noise. By maximizing coding rate reduction during training, ERASE adapts to diverse data distributions, improving robustness against label noise.

\section{\hspace{-0.1cm}Training-Time Graph OOD Adaptation}
\label{sec:TrTA}

Training-time graph OOD adaptation 
is a crucial aspect of addressing distribution discrepancies between source and target graph data.
These discrepancies can arise due to various factors, including differences in the underlying graph structures, node features, and edge relationships. Effectively mitigating these distribution shifts is essential for enabling the model to transfer knowledge from the source domain and adapt to the target domain.
In general, training-time graph OOD adaptation serves three primary objectives in different settings.
\begin{itemize}[wide=0pt]
    \item \textit{Observation Bias Correction.} For semi-supervised classification within a single graph, distribution shifts between training and test instances may arise from observation bias related to latent subpopulation \cite{Bi2023PredictingTS}, or the time-evolving nature of graphs \cite{Zhu2022ShiftRobustNC}. 
    Mitigating distribution shifts in this setting may enhance the model's performance on test instances.
    \item \textit{Cross-graph Knowledge Transfer.}
    Cross-graph knowledge transfer is a critical challenge when transferring knowledge from well-labeled graphs to graphs with limited or no labels. Distinct graphs often exhibit varied data distributions, leading to discrepancies in the representations learned from each graph. This objective ensures that knowledge learned from the source graph can be effectively applied to the target graph.
    \item \textit{Negative Augmentation Mitigation.}
    Graph data augmentation, which utilizes the augmented data as additional training data, is commonly used for improving model generalization or alleviating label scarcity issues. However, overly severe distribution shifts between the original and augmented data can lead to the negative augmentation problem \cite{wu2022knowledge, liu2022confidence}. 
    Therefore, controlling distribution shifts is essential to avoid inferior model performance and fully exploit the benefits of augmented graph data. 
\end{itemize}

In this section, we discuss existing training-time graph OOD adaptation methods, highlighting the techniques for mitigating distribution shifts.
The summary can be found in Table \ref{tab:1}.

\subsection{Model-Centric Approaches}
Model-centric methods for training-time graph OOD adaptation
can be further categorized into \textit{invariant representation learning} and \textit{concept-shift aware representation learning}, which aim at learning distributionally aligned representations, and \textit{model regularization}, which focuses on achieving effective knowledge transfer through model regularization.

\smallskip 
\noindent 
\textbf{Invariant Representation Learning.} 
Unlike the invariant representation learning applied in graph OOD generalization, which focuses on generalizing to unseen target distributions without any access to target domain data, invariant representation learning in training-time graph OOD adaptation explicitly considers both source and target domain data during training. This allows the model to align representations across both domains while addressing distribution shifts.
This technique is frequently employed for domain adaptation under covariate assumption\footnote{Within the context of graph OOD adaptation, we commonly use the terms ``distribution'' and ``domain'' interchangeably.}, in which an invariant relationship between latent representations and labels $\mathcal{P}_S(\textbf{Y}|\textbf{H})=\mathcal{P}_T(\textbf{Y}|\textbf{H})$ is assumed.
Inspired by the theoretical generalization bound \cite{ben2006analysis}, domain-invariant representation learning methods aim to train a graph encoder $g(\cdot)$ 
to minimize discrepancies between the induced marginal source distribution $\mathcal{P}_S(\textbf{H})$ and the target distribution $\mathcal{P}_T(\textbf{H})$, while also identifying a classifier $f(\cdot)$ in the latent space that minimizes empirical source risk.
To achieve these two goals, the loss function for domain-invariant representation learning can be formulated as follows:
\begin{equation}
\label{eq:1}
\min_{f,g} \mathbb{E}_{\textbf{A, X, Y}} [l(f ( g (\textbf{A}, \textbf{X})),  \textbf{Y} ) ] + l_{reg},
\end{equation}
where $l_{reg}$ denotes a regularization term that
facilitates the alignment of the induced marginal distribution $\mathcal{P}(\textbf{H})$. 
Three strategies are mainly adopted:
\textit{(1) distribution distance minimization}, \textit{(2) adversarial learning}, and \textit{(3) disentangled representation learning}.

    \begin{itemize}[wide]
    \item \textit{(1) Distribution Distance Minimization.}
    Distribution distance minimization directly employs the distance between marginal distributions as the regularization term in Equation \ref{eq:1}.
    Different methods vary in their choice of distance metrics and the specific representations they aim to align. SR-GNN \cite{Zhu2021ShiftRobustGO} considers central moment discrepancy (CMD) as regularization and aligns the distribution discrepancies in the final layer of traditional GCN.
    CDNE \cite{Shen2019NetworkTN}, GraphAE \cite{Guo2023LearningAN}, and GRADE \cite{Wu2022NonIIDTL} target at minimizing the statistical discrepancies between source and target across all latent layers, with the regularization term as a summation of distribution distances of different layers.
    Regarding the choice of distance metric, CDNE utilizes marginal maximum mean discrepancy and class-conditional marginal maximum mean discrepancy. GraphAE adopts the multiple-kernel variant of maximum mean discrepancy, while GRADE defines and applies subtree discrepancy. 
    JHGDA \cite{Shi2023ImprovingGD} relies on a hierarchical pooling module to extract network hierarchies and minimizes statistical discrepancies in hierarchical representations via the exponential form of the marginal and class-conditional maximum mean discrepancy.
    For non-trainable representations, such as the latent embeddings in SimpleGCN \cite{wu2019simplifying}, SR-GNN \cite{Zhu2021ShiftRobustGO} applies an instance weighting technique. In this approach, the learnable weight parameters are optimized using kernel mean matching to mitigate distribution discrepancies. HC-GST \cite{wang2024hc} uses pseudo-labels, incorporating heterophily ratios and labeling accuracy to align distributions, with learnable weights optimized via kernel mean matching. Most recently, DREAM \cite{yin2024dream} employs complementary branches (graph-level and subgraph-enhanced) and enforces consistency between them to mitigate biased pseudo-labels and reduce distribution discrepancies. Addtionally, SelMAG~\cite{zhao2024multi} designs an optimal-transport-based algorithm by selecting informative source graphs and adapting both embedding and classification spaces. It further employs meta-learning to refine subgraph and node selection, improving transferability.
    \item \textit{(2) Adversarial Learning.} Adversarial learning approaches align representations by training the graph encoder
    $g(\cdot)$ to generate embeddings that confuse the domain discriminator $f_d(\cdot)$. Here, the domain discriminator is trained to differentiate between source and target domain representations, while the graph encoder tries to produce embeddings that make this distinction difficult.
    Correspondingly, the regularization term $l_{reg}$ in Equation \ref{eq:1} is usually formulated as a minimax game between $g(\cdot)$ and $f_d(\cdot)$ as:

    \begin{equation}
    \label{eq:6}
    \min_{f_d} \max_{g} l(f_d(g(\textbf{A}, \textbf{X})), \textbf{Y}_d),
    \end{equation}    
    where $\textbf{Y}_d$ denotes the domain label, and $l(\cdot)$ can be chosen as a negative distance loss \cite{Dai2019GraphTL}, or a domain classification loss \cite{Wu2019DomainAdversarialGN,Wu2020UnsupervisedDA,Shen2020AdversarialDN,ding2021cross,Guo2023LearningAN,Qiao2023SemisupervisedDA,wang2024open}.
    In addition to framing it as a minimax problem, DANE \cite{Zhang2019DANEDA} explores the use of two symmetric and adversarial losses for model training, aiming to achieve bi-directional transfer.
    Typically, adversarial alignment takes place in the final hidden layer, with the exception being GraphAE \cite{Guo2023LearningAN}, which aligns representations in all hidden layers.
    ASN \cite{Zhang2021AdversarialSN} introduces a decomposition of the learned representation into two components: a domain-private component and a domain-invariant component related to classification. A domain adversarial loss is additionally added to facilitate the learning of invariant representations.
    In addition, it is noteworthy that SGDA \cite{Qiao2023SemisupervisedDA} also takes the label scarcity issue of the source graph into account by employing a weighted self-supervised pseudo-labeling loss. SDA \cite{wang2024open} studies the problem under the open-set setting, utilizing an entropy-based separation strategy to categorize target nodes into certain and uncertain groups and specifically aligning nodes from the certain group with adversarial learning technique. Most recently, JDA-GCN~\cite{ijcai2024p276} introduces a joint adversarial domain adaptive graph convolutional network, leveraging both local and global graph structures through structural graph alignment, improving the model's ability to capture intricate dependencies within graph data.

    \item \textit{(3) Disentangled Representation Learning.}
    In training-time graph OOD adaptation, the focus is on learning representations that not only separate domain-invariant features but also adapt to the specific distribution shifts between source and target domains. Disentangled representation learning separates representations into distinct, informative factors. Among these factors, one is designed to capture domain-invariant features that are essential for the target task (e.g classification). This domain-invariant component remains consistent across different domains to preserve classification-related semantic information, thereby facilitating effective adaptation. 
    And the disentangled components can be selectively manipulated to cater to specific requirements of the target task.
    The loss function for disentangled representation learning can be defined as:
    \begin{equation}
    \label{eq:7}
    \min_{f, g_s} \mathbb{E}_{\textbf{A, X, Y}} [l(f ( g_s (\textbf{A}, \textbf{X})), \textbf{Y})] +  \min_{g_s, g_o} (l_{reg} + l_{recon} + l_{add}),
    \end{equation}
    where $g_s$ 
    denotes the graph encoder
    for acquiring invariant task-related information, while 
    $g_o$ refers to encoders for other components excluding $g_s$.
    The term $l_{reg}$ represents a regularization term designed to enhance the separation between different components, $l_{recon}$ is the reconstruction loss used to recover the original graph structure from the concatenated representation, preventing information loss, and $l_{add}$ introduces additional terms that facilitate learning disentangled representations, ensuring that specific components exhibit the desired characteristics.
    Analogous to DIVA \cite{Ilse2019DIVADI}, DGDA \cite{Cai2021GraphDA} assumes that the graph generation process is controlled independently by domain-invariant semantic latent variables, domain latent variables, and random latent variables. 
    To learn representations with desired characteristics, domain classification loss and noise reconstruction loss are considered as the additional losses.
    \end{itemize}

\smallskip 
\noindent
\textbf{Concept-shift Aware Representation Learning.}
Some recent works extend beyond the covariate assumption and take the change of label function across domains into consideration.
As illustrated in \cite{zhao2019learning},
when there exist concept shifts in $\mathcal{P}(\textbf{Y}|\textbf{X})$ or $\mathcal{P}(\textbf{Y}|\textbf{H})$, namely, the label function changes, the inestimable adaptability term in the upper bound \cite{ben2006analysis} may be large and the performance of previous invariant representation learning methods on target is no longer guaranteed.
A similar upper bound and example provided in~\cite{Liu2023StructuralRI} illustrate the insufficiency of invariant representation learning methods based on the covariate shift assumption.

To further accommodate the change in label function, 
SRNC~\cite{Zhu2022ShiftRobustNC} leverages graph homophily, incorporating a shift-robust classification GNN module and an unsupervised clustering GNN module to alleviate the distribution shifts in the joint distribution $\mathcal{P}(\textbf{H}, \textbf{Y})$.
Notably, SRNC is also capable of handling the open-set setting where new classes emerge in the test data.
In StruRW~\cite{Liu2023StructuralRI}, 
the conditional structure shifts $\mathcal{P}(\textbf{A}|\textbf{Y})$ are identified and mitigated by adaptively adjusting the weights of edges in the source graph. 
Building on this, Pair-Align~\cite{liu2024pairwise} is proposed as an extension of StruRW to address both conditional structure shift and label shift.
Moreover, Zhu et al.~\cite{Zhu2023ExplainingAA} demonstrate that, in the contextual stochastic block model, conditional shifts in the latent space $\mathcal{P}(\textbf{Y}|\textbf{H})$ can be exacerbated by both graph heterophily and graph convolution in the GCN compared to conditional shifts in input feature space $\mathcal{P}(\textbf{Y}|\textbf{X})$. 
Hence, they introduce GCONDA that explicitly matches the distribution of $\mathcal{P}(\textbf{Y}|\textbf{H})$ across domains via Wasserstein distance regularization, and additionally, they also propose GCONDA++ that jointly minimizes the discrepancy in $\mathcal{P}(\textbf{Y}|\textbf{H})$ and $\mathcal{P}(\textbf{H})$. 

\smallskip 
\noindent
\textbf{Model Regularization.}
Instead of focusing on the process of learning aligned representations, some other methods achieve effective knowledge transfer under distribution shifts through model regularization strategy, including spectral regularization \cite{You2023GraphDA, liu2024rethinking} and knowledge distillation \cite{wu2022knowledge, Bi2023PredictingTS}. 
Building on the derived GNN-based generalization bound, You et al.~\cite{You2023GraphDA} propose SSReg and MFRReg, which regularize the spectral properties of GNN to enhance transferability.
They also extend their theoretical results to the semi-supervised setting with the challenging distribution shifts in $\mathcal{P}(\textbf{Y}|\textbf{A}, \textbf{X})$.
Liu et al.~\cite{liu2024rethinking} then 
delve into the inherent generalization ability of GNN models, suggesting the removal of propagation layers in the source graph while stacking multiple propagation layers in the target graph. 
This design inherently acts as a regularization for the GNN Lipschitz constant, leading to a tighter generalization bound.
Both KDGA~\cite{wu2022knowledge} and KTGNN \cite{Bi2023PredictingTS} employ knowledge distillation, enforcing Kullback–Leibler (KL) divergence regularization between the outputs of teacher and student models.
KDGA specifically addresses the negative augmentation problem by distilling knowledge from a teacher model trained on augmented graphs into a partially parameter-shared student model trained on the original graph. 
In contrast, KTGNN focuses on semi-supervised node classification in VS-Graph, where vocal nodes serve as the source and silent nodes with incomplete features act as the target. Separate classifiers are constructed for the source and target domains, and their knowledge is distilled into a transferable student classifier using KL regularization.

\begin{table*}[ht]
\setlength\tabcolsep{5.pt}
\centering
\caption{A summary of training-time and test-time graph OOD adaptation methods.}
\scalebox{0.75}{
\begin{tabular}{c|c|cccccc}
\bottomrule
\rowcolor{Gray} & &  &  &  & & &  \\
\rowcolor{Gray}  
\multirow{-2}{*}{\textbf{Scenario}} & \multirow{-2}{*}{\textbf{Approach}} &\multirow{-2}{*}{\textbf{Technique}} & \multirow{-2}{*}{\textbf{Method}}  & \multirow{-2}{*}{\makecell{\textbf{Task} \\ \textbf{Level}}} & \multirow{-2}{*}{\makecell{\textbf{Domain} \\ \textbf{Source}}} & \multirow{-2}{*}{\makecell{\textbf{Distribution} \\ \textbf{Shift}}}& \multirow{-2}{*}{\textbf{Objective}} \\ \toprule
\multirow{50}{*}{\rotatebox{90}{\makecell{Training-time Graph OOD Adaptation}}} &
\multirow{36}{*}{\rotatebox{90}{\makecell{Model-centric}}} &
\multirow{25}{*}{\makecell{Invariant\\Representation\\Learning}} 
& DAGNN \cite{Wu2019DomainAdversarialGN} & Graph & Single & Covariate & Cross-graph transfer   \\  \cmidrule{4-8}
& & & DANE \cite{Zhang2019DANEDA} & Node & Single & Covariate & Cross-graph transfer   \\  \cmidrule{4-8}
& & & CDNE \cite{Shen2019NetworkTN} & Node & Single & Covariate\&Concept & Cross-graph transfer   \\  \cmidrule{4-8}
& & & ACDNE \cite{Shen2020AdversarialDN} & Node & Single & Covariate & Cross-graph transfer   \\  \cmidrule{4-8}
& & & UDA-GCN \cite{Wu2020UnsupervisedDA} & Node & Single & Covariate & Cross-graph transfer   \\  \cmidrule{4-8}
& &  & DGDA \cite{Cai2021GraphDA} & Graph & Single & Covariate & Cross-graph transfer   \\ \cmidrule{4-8}
& & & SR-GNN \cite{Zhu2021ShiftRobustGO} & Node & Single & Covariate & Observation bias correction   \\ \cmidrule{4-8}
& & & COMMANDER \cite{ding2021cross} & Graph & Single & Covariate & Cross-graph transfer  \\ \cmidrule{4-8}
& & & ASN \cite{Zhang2021AdversarialSN} & Node & Single & Covariate & Cross-graph transfer   \\  \cmidrule{4-8}
& &  & AdaGCN \cite{Dai2019GraphTL} & Node & Single & Covariate\&Concept & Cross-graph transfer   \\  \cmidrule{4-8}
& &  & GraphAE \cite{Guo2023LearningAN} & Node & Single/Multiple & Covariate& Cross-graph transfer   \\  \cmidrule{4-8}
& &  & GRADE \cite{Wu2022NonIIDTL} & Node/Edge & Single & Covariate & Cross-graph transfer   \\  \cmidrule{4-8}
& &  & JHGDA \cite{Shi2023ImprovingGD} & Node & Single & Covariate & Cross-graph transfer   \\  \cmidrule{4-8}
& &  & SGDA \cite{Qiao2023SemisupervisedDA} & Node & Single & Covariate & Cross-graph transfer   \\  \cmidrule{4-8}
& &  & SDA \cite{wang2024open} & Node & Single & Covariate\&Concept & Cross-graph transfer \\  \cmidrule{4-8}
& &  & JDA-GCN \cite{ijcai2024p276} & Node & Single & Covariate & Cross-graph transfer \\  \cmidrule{4-8}
& &  & HC-GST \cite{wang2024hc} & Node & Single & Covariate & Observation bias correction \\  \cmidrule{4-8}
& &  &  DREAM~\cite{yin2024dream} & Graph & Single& Concept & Cross-graph transfer \\
\cmidrule{4-8}
& & & SelMAG~\cite{zhao2024multi} & Node & Multiple & Covariate\&Concept & Cross-graph transfer  \\  \cmidrule{3-8}
& & \multirow{5}{*}{\makecell{Concept-shift\\ Aware\\ 
Representation\\ Learning}} & SRNC \cite{Zhu2022ShiftRobustNC} & Node & Single & Covariate\&Concept & Cross-graph transfer/Observation bias correction   \\ \cmidrule{4-8}
& & & StruRW \cite{Liu2023StructuralRI} & Node & Single/Multi & Concept (Structure) & Cross-graph transfer   \\ \cmidrule{4-8}
& & & Pair-Align \cite{liu2024pairwise} & Node & Single & Covariate\&Concept & Cross-graph transfer   \\ \cmidrule{4-8}
& & & GCONDA++ \cite{Zhu2023ExplainingAA} & Node/Graph &Single & Covariate\&Concept & Cross-graph transfer/Observation bias correction   \\  \cmidrule{3-8}
& & \multirow{4}{*}{\makecell{Model \\Regularization}} 
& KDGA \cite{wu2022knowledge} & Node & Multiple & Concept (Structure) & Negative augmentation mitigation  \\ \cmidrule{4-8}
& &  & SS/MFR-Reg \cite{You2023GraphDA} & Node/Edge & Single & Covariate\&Concept & Cross-graph transfer    \\ 
\cmidrule{4-8}
& &  & A2GNN \cite{liu2024rethinking} & Node & Single & Covariate    & Cross-graph transfer    \\  \cmidrule{4-8}
& &  & KTGNN \cite{Bi2023PredictingTS} & Node & Single & Concept (Feature) & Observation bias correction   \\ \cmidrule{2-8}
& \multirow{11}{*}{\rotatebox{90}{\makecell{Data-centric}}} &\multirow{4}{*}{\makecell{Instance\\ weighting}} & IW \cite{Ye2013PredictingPA} & edge & Single & Concept (Struture) & Cross-graph transfer   \\ \cmidrule{4-8}
& & & NES-TL \cite{Fu2020NESTLNE} & Node & Multiple & Concept (Struture) & Cross-graph transfer   \\ \cmidrule{4-8}
& & & RSS-GCN \cite{Wu2022ReinforcedSS} & Graph & Multiple &Covariate & Cross-graph transfer   \\ \cmidrule{4-8}
& & & DR-GST \cite{liu2022confidence} & Node & Single & Covariate\&Concept & Negative augmentation mitigation  \\  \cmidrule{3-8}
& &\multirow{5}{*}{\makecell{Graph Data\\Augmentation}}   
& FakeEdge \cite{Dong2022FakeEdgeAD} & Edge & Single & Concept (Structure) & Observation bias correction   \\ \cmidrule{4-8}
& & & Bridged-GNN \cite{Bi2023BridgedGNNKB} & Node & Single & Covariate\&Concept & Cross-graph transfer   \\ \cmidrule{4-8}
&  && DC-GST \cite{wang2024distribution} & Node & Multiple & Covariate & 
Negative augmentation mitigation/Bias correction   \\ \cmidrule{4-8}
& & & LTLP \cite{wang2024optimizing} & Edge & Single & Covariate & Negative augmentation mitigation/Cross-graph transfer  \\
\midrule
\multirow{16}{*}{\rotatebox{90}{\makecell{Test-time Graph OOD Adaptation}}} &
\multirow{12}{*}{\rotatebox{90}{\makecell{Model-centric}}} & \multirow{5}{*}{\makecell{Semi-supervised\\Fine-tuning}} & 
GraphControl \cite{Zhu2023GraphControlAC} & Node & Multiple & Concept (Feature) & Avoid over-fitting   \\ \cmidrule{4-8}
 & & & G-Adapter \cite{gui2024g} & Graph & Single & Covariate\&Concept & Avoid over-fitting   \\ \cmidrule{4-8}
& & & AdapterGNN \cite{li2024adaptergnn} & Graph & Single & Covariate\&Concept & Avoid over-fitting    \\  \cmidrule{4-8}
& & & PROGRAM \cite{sun2024program} & Node & Multiple & Covariate\&Concept & Avoid over-fitting  \\  \cmidrule{3-8}
& & \multirow{3}{*}{\makecell{Self-supervised\\ Adaptation}} 
& SOGA \cite{Mao2021SourceFU} & Node & Single & Covariate & Maintain discriminative ability   \\ \cmidrule{4-8}
& &  & GAPGC \cite{Chen2022GraphTTATT} & Graph & Single & Covariate & Maintain discriminative ability    \\ 
\cmidrule{4-8}
& &  & GT3 \cite{Wang2022TestTimeTF}  & Graph& Multiple & Covariate & Maintain discriminative ability    \\ \cmidrule{2-8}
& \multirow{4}{*}{\rotatebox{90}{\makecell{Data-\\centric}}} & \multirow{1}{*}{\makecell{Feature Reconstruction}}
& FRGNN \cite{Ding2023FRGNNMT}  & Node & Single & Concept (Feature) & Mitigate bias   \\ \cmidrule{3-8}
& & \multirow{2}{*}{\makecell{Graph Data\\Augmentation}}  
& GTRANS \cite{Jin2022EmpoweringGR} & Node & Single & Covariate & Maintain discriminative ability    \\ \cmidrule{4-8}
& & & GraphCTA \cite{zhang2024collaborate} & Node & Single & Covariate & Maintain discriminative ability   \\ 
\bottomrule
\end{tabular}
}
\label{tab:1}
\end{table*}

\subsection{Data-Centric Approaches}
Data-centric approaches for training-time graph OOD adaptation leverage the unique structure and properties of graph data to address distribution shifts. These methods enhance the graph’s input data by considering its inherent topological and relational features, which are crucial for effective adaptation across domains. Two prominent techniques in this context are instance weighting and graph data augmentation, both designed to mitigate distribution shifts and improve model performance.

\smallskip
\noindent
\textbf{Instance Weighting.} 
Instance Weighting, which assigns different weights to data points, is a commonly used data-centric technique in traditional transfer learning \cite{Zhuang2020ACS}. 
Similar strategies have been applied to graph-based OOD adaptation, where instance weighting methods include edge weighting \cite{Ye2013PredictingPA}, node weighting \cite{liu2022confidence}, and graph weighting \cite{Fu2020NESTLNE, Wu2022ReinforcedSS}, each designed for specific tasks.
Inspired by Adaboost and TrAda,
Ye et al.~\cite{Ye2013PredictingPA} employ instance weighting for the edge sign prediction task. They iteratively adjust edge weights, decreasing the weights of misclassified dissimilar source instances to mitigate distribution shifts across graphs. 
On the other hand, 
Liu et al.~\cite{liu2022confidence} recognize that the distribution shifts between the original data and the augmented data with pseudo-labels may impede the effectiveness of self-training, and thus propose to reweight the augmented node instances based on information gain to mitigate the gap between the original distribution and the shifted distribution.

In multi-source transfer settings, not all source graphs contribute equally to predictions on the target graph, and some may be of low quality. To address this, both RSS-GCN \cite{Wu2022ReinforcedSS} and NES-TL \cite{Fu2020NESTLNE} employ graph weighting techniques to effectively integrate multiple source graphs.
NES-TL \cite{Fu2020NESTLNE} proposes the NES index to quantitatively measure the structural similarity between two graphs, and uses the NES-based scores as weights to ensemble weak classifiers trained on instances from each source graph along with labeled target instances.
Additionally, RSS-GCN \cite{Wu2022ReinforcedSS} utilizes reinforcement learning to select high-quality source graphs, aiming to minimize the distribution divergence between selected source and target graphs. This sample selection strategy can be considered as a special case of binary instance weighting.

\smallskip
\noindent
\textbf{Graph Data Augmentation.} 
Unlike the instance weighting strategy, which assigns weights to graph components, graph data augmentation methods modify graph structures to mitigate distribution shifts. It is important to clarify that this discussion focuses on graph data augmentation for training-time adaptation rather than generalization, as the former specifically aims to address distribution shifts during training, mainly focusing on edge adaptation. 
Dong et al.~\cite{Dong2022FakeEdgeAD} identify that distribution shifts in edge prediction stem from the presence of links in training but their absence during testing.
To address this, they propose FakeEdge, a subgraph-based link prediction framework that intentionally adds or removes the focal links within the subgraph.
This method separates the dual role of edges, treating them both as features for representation learning and as labels for link prediction.
Bi et al.~\cite{Bi2023BridgedGNNKB} 
approach the domain-level knowledge transfer problem by redefining it as learning a sample-wise knowledge-enhanced posterior distribution. In their framework, they compute similarities between source and target samples, constructing bridges that connect each sample to its similar samples containing valuable knowledge for prediction.
A GNN model is then trained to transfer knowledge across source and target samples using this bridged-graph.
This approach allows for smoother adaptation by explicitly bridging domain discrepancies between the source and target graphs, ensuring that knowledge from the source domain is transferred to the target domain effectively.
More recently, a novel framework called DC-GST \cite{wang2024distribution} has been introduced to bridge the distribution shifts between augmented training instances and test instances in self-training, which incorporates a distribution-shift-aware edge predictor to improve the model's generalizability of assigning pseudo-labels, along with a carefully designed criterion for pseudo-label selection.
Additionally, LTLP~\cite{wang2024optimizing} enhances graph data augmentation by addressing the long-tailed distribution of common neighbors in link prediction. 
It achieves this by generating high-quality edges to increase the number of common neighbors for tail node pairs and refining their representations.

\section{Test-Time Graph OOD Adaptation}
\label{sec:TTA}
\smallskip 
Test-time graph OOD adaptation focuses on adjusting a pre-trained model to new target data that becomes available after the initial training phase. Unlike training-time adaptation, which requires simultaneous access to both source and target data, test-time adaptation updates the model using only the target data.
This approach, often referred to as source-free adaptation, is particularly important in scenarios where access to source data is restricted. For instance, in social networks, source data is typically confidential and inaccessible due to privacy protection regulations and concerns about data leakage.
In this section, we delve into test-time graph OOD adaptation. 
A summary for related models is provided in Table \ref{tab:1}.

\subsection{Model-Centric Approaches}
Fine-tuning is a prevalent method for adapting a pre-trained model during test time, as it allows the model to leverage knowledge learned from large-scale datasets. However, effectively utilizing information from the pre-trained model presents challenges, which are typically addressed through the following techniques.

\smallskip 
\noindent
\textbf{Semi-supervised Fine-tuning.} 
This approach generally involves pre-training the model in an unsupervised manner to capture more transferable structural information. During fine-tuning, task-specific and domain-related information is incorporated to enhance adaptation.
A notable challenge here is the scarcity of labels in the target graphs, which can lead to overfitting.
To avoid negative transfer, both G-Adapter \cite{gui2024g} and AdapterGNN \cite{li2024adaptergnn} employ the parameter-efficient fine-tuning (PEFT) strategy that takes bottleneck structure to alleviate overfitting by reducing the size of the tunable parameter.
Alternatively, GraphControl \cite{Zhu2023GraphControlAC} introduces an adaptive mechanism for integrating target information. It consists of two components: a frozen pre-trained model that takes the adjacency matrix as input and a trainable copy that takes a node feature-based kernel matrix as input. These components are linked through zero MLPs with expanding parameters, designed to reduce noise in target node features while gradually incorporating downstream information into the pre-trained model.
PROGRAM~\cite{sun2024program} builds robust connections between test data and prototypes with two key components: a prototype graph model that generates reliable pseudo-labels by leveraging relationships between prototypes and test data, and a robust self-training module that iteratively refines these pseudo-labels  through consistency regularization.

\smallskip 
\noindent
\textbf{Self-supervised Adaptation.} 
In this technique, task-related information is first encoded into the pre-trained model, followed by an unsupervised fine-tuning task on the target graph.
However, a major challenge arises during this unsupervised phase: the model may lose its discriminative ability for the primary task or potentially learn irrelevant information. To address this issue, various techniques have been proposed to retain the model's discriminative power.
SOGA \cite{Mao2021SourceFU} and GAPGC \cite{Chen2022GraphTTATT} employ information-based design strategies while GT3 \cite{Wang2022TestTimeTF} tackles the challenge by intentional architecture design and additional regularization constraints.
SOGA, employs a loss function that maximizes the mutual information between the inputs and outputs of the model to enhance the discriminatory power. GAPGC \cite{Chen2022GraphTTATT} employs an adversarial pseudo-group contrast strategy to tackle over-confidence bias and mitigate the risk of capturing redundant information, which has a lower bound guarantee of the information relevant to the main task from the information bottleneck perspective.
On the other hand, GT3 \cite{Wang2022TestTimeTF} structures the model to include two branches that share initial layers: a main task (classification) branch and a self-supervised branchDuring test time, only the self-supervised branch is fine-tuned, ensuring that the discriminative ability of the main-task branch is preserved. Additionally, GT3 integrates additional regularization constraints between training and test-time output embeddings, hence enforcing their statistical similarity and avoiding substantial deviations.

Other relevant studies, such as GTOT-Tuning \cite{Zhang2022FineTuningGN}, which focuses on transferring knowledge across tasks, and GraphGLOW \cite{Zhao2023GraphGLOWUA}, which incorporates sharable components directly into the model, also provide valuable insights into addressing distribution shifts during test-time adaptation.

\subsection{Data-Centric Approaches}

Data-centric approaches in test-time graph OOD adaptation focus on modifying input data rather than altering the pre-trained model. Unlike generalization techniques, which aim to improve robustness across diverse distributions, and training-time adaptation, which requires access to both source and target data, test-time adaptation operates solely on target data. These methods emphasize the reconstruction and augmentation of graph features to enhance model performance without modifying the pre-trained architecture.

\smallskip 
\noindent
\textbf{Feature Reconstruction.}
Feature reconstruction aims to mitigate distribution shifts between training and target through reconstructing node features of target graphs to resemble those of source graphs during test time.
In FRGNN, 
Ding et al.~\cite{Ding2023FRGNNMT} consider the semi-supervised node classification task and utilize an MLP to establish a mapping between the output and input space of the pre-trained GNN. Subsequently, using the encoded one-hot class vectors as inputs, the MLP generates class representative representations. 
By substituting the features of the labeled test nodes with the representative representations of the corresponding classes and spreading the updated information to other unlabeled test nodes through message passing, the graph embedding bias between test nodes and training nodes is anticipated to be mitigated.

\smallskip 
\noindent
\textbf{Graph Data Augmentation.}
Apart from reconstructing features, Jin et al.~\cite{Jin2022EmpoweringGR} introduce a graph transformation framework called GTRANS to address distribution shifts during test time. 
This framework models graph transformation as the injection of perturbations into both graph structures and node features, optimizing these transformations through a parameter-free surrogate loss.
Theoretical analyses are also provided to guide the selection of appropriate surrogate loss functions.
Building upon GTRANS, GraphCTA \cite{zhang2024collaborate} further integrates model adaptation and graph adaptation in a collaborative manner. 
The collaborative loop is formed by adapting the model using predictions from the node’s neighborhood, enhancing the graph through neighborhood contrastive learning, and then reintroducing the adapted graph into the model adaptation process. 
It is worth highlighting that such data-centric test-time graph OOD adaptation methods that prioritize adjusting the test data over modifying the pre-trained model can be especially beneficial when handling large-scale pre-trained models. 

\section{Dataset}
\label{sec:dataset}
\begin{table*}[ht]
\setlength\tabcolsep{7.pt}
\centering
\caption{A summary of graph OOD datasets. "Avg. Nodes" and "Avg. Edges" represent the average number of nodes and edges in each graph, respectively. "Classes" indicates the number of target categories for classification tasks. "Graphs" specifies the total number of graphs in each dataset. "Scenario" refers to the primary context of generalization (G.) or adaptation (A.), while "Shift Types" denote the types of distribution shifts, with covariate (Cov.) and concept (Con.) shifts. "Domain Separation" identifies the factors that separate the domains. "Field" describes the nature of the data, and "Task" highlights the level at which the dataset can be used.}
\scalebox{0.9}{
\begin{tabular}{cccccccccc}
\bottomrule
\rowcolor{Gray} & & & & & & & & & \\
\rowcolor{Gray}  
\multirow{-2}{*}{\textbf{Dataset}} & \multirow{-2}{*}{\makecell{\textbf{Avg.} \\ \textbf{Nodes}}} & \multirow{-2}{*}{\makecell{\textbf{Avg.} \\ \textbf{Edges}}} & \multirow{-2}{*}{\makecell{\textbf{Classes}}} & \multirow{-2}{*}{\makecell{\textbf{Graphs}}} & \multirow{-2}{*}{\makecell{\textbf{Scenarios}}} & \multirow{-2}{*}{\makecell{\textbf{Shift} \\ \textbf{Types}}} & \multirow{-2}{*}{\makecell{\textbf{Domain} \\ \textbf{Separation}}} & \multirow{-2}{*}{\makecell{\textbf{Field}}} & \multirow{-2}{*}{\makecell{\textbf{Task} \\ \textbf{Level}}} \\ \toprule
Spurious-Motif~\cite{wu2022discovering} & 44.96 & 65.67 & 3 & 18000 & G. & Cov.\&Con. & Size & Synthetic graph & Graph \\
MNIST-75sp~\cite{knyazev2019understanding} & 70.97 & 590.52 & 10 & 35000 & G. & Cov.\&Con. & Feature noise & Superpixel graph & Graph \\
Graph-SST2~\cite{yuan2023explainability} & 10.2 & 18.4 & 2 & 43689 & G. & Cov. & Node degree & Text sentiment & Graph \\
DrugOOD-Assay~\cite{ji2022drugood} & 32.27 & 70.25 & 2 & 72239 & G. & Cov. & Assay & Molecular graph & Graph \\
DrugOOD-Scaffold~\cite{ji2022drugood} & 29.95 & 64.86 & 2 & 59608 & G. & Cov. & Scaffold & Molecular graph & Graph \\
DrugOOD-Size~\cite{ji2022drugood} & 30.73 & 66.90 & 2 & 70672 & G. & Cov. & Size & Molecular graph & Graph \\
D\&D~\cite{morris2020tudataset} & 284.32 & 1431.32 & 2 & 1178 & G. & Cov. & Graph size & Molecular graph & Graph \\
NCI1~\cite{morris2020tudataset} & 29.87 & 32.30 & 2 & 4110 & G. & Cov. & Graph size & Molecular graph & Graph \\
OGBG-MolPCBA~\cite{hu2020open} & 26.0 & 28.1 & 2 & 437929 & G. & Cov.\&Con. & Scaffold/Size & Molecular graph & Graph \\
OGBG-MolHIV~\cite{hu2020open} & 25.5 & 27.5 & 2 & 41127 & G. & Cov.\&Con. & Scaffold/Size & Molecular graph & Graph \\
IMDB~\cite{yanardag2015deep} & 19.77 & 96.53 & 2 & 1000 & G. & Cov. & Time & Collaboration network & Graph \\
Twitter~\cite{qiu2018deepinf} & 0.91 & 2.51 & 2 & 499160 & A. & Cov. & Time & Social network & Graph \\
Reddit-Binary~\cite{hamilton2017inductive} &429.63 &  497.75 & 2 & 2000 & G./A. & Con. & Community & Social network & Graph/Node  \\
OGBG-PPA~\cite{hu2020open} & 243.4 & 2266.1 & 37 & 158110 & G. & Cov.\&Con. & Species & Protein network & Graph \\
OGBN-Proteins~\cite{hu2020open}  & 132534 & 39561252 & 2 & 1 & G. & Cov. & Species & Protein network & Node \\
PPI~\cite{hamilton2017inductive} & 2372.7 & 68508.7 & 24 & 24 & G. & Cov.\&Con. & Graph & Protein network & Node/Edge \\
OGBN-Products~\cite{hu2020open} & 2449029 & 61859140 & 47 & 1 & G. & Cov. & Popularity & Co-purchase network & Node \\
Amazon-Photo~\cite{shchur2018pitfalls} & 7650 & 238162 & 8 & 1 & G./A. & Cov. & Graph & Co-purchase network & Node \\
Cora~\cite{bojchevski2018deep} & 2708 & 5429 & 7 & 1 & G./A. & Cov.\&Con. & Word/Degree & Citation network & Node \\
Citeseer~\cite{giles1998citeseer} & 3327 & 4732 & 6 & 1 & G./A. & Cov.\&Con. & Word/Degree & Citation network & Node \\
Pubmed~\cite{namata2012query} & 19717 & 44338 & 3 & 1 & G./A. & Cov.\&Con. & Word/Degree & Citation network & Node \\
OGBN-Arxiv~\cite{hu2020open} & 169343 & 1166243 & 40 & 1 & G./A. & Cov.\&Con. & Time & Citation network & Node \\
DBLP~\cite{Zhu2023ExplainingAA} & 39254.5 & 500650 & 5 & 2 & G./A. & Cov.\&Con. & Time & Citation network & Node \\
ACM~\cite{Zhu2023ExplainingAA} & 11671.5 & 81053 & 5 & 2 & G./A. & Cov.\&Con. & Time & Citation network & Node \\
Yelp~\cite{zhao2024multi} & 2317.17 & 38754.17 & 6 & 6 & G./A. & Cov.\&Con. & City & Social network & Node/Edge \\
Facebook100~\cite{traud2012social} & 12066.08 & 469028.26 & 2 & 100 & G. & Cov.\&Con. & University & Social network & Node \\
Twitch~\cite{rozemberczki2021multi} & 5686.67 & 148724.33 & 2 & 6 & G. & Cov.\&Con. & Language & Gamer network & Node  \\
Elliptic~\cite{pareja2020evolvegcn} & 203769 & 234355 & 2 & 1 & G./A. & Con. & Time & Bitcoin Transactions & Node  \\
GOOD-WebKB~\cite{gui2022good} & 617 & 1138 & 5 & 1 & G. & Cov.\&Con. & University & Webpage network & Node \\
BlogCatalog~\cite{guo2020learning} & 5196 & 171743 & 6 & 2 & A. & Cov.\&Con. &  Group affiliation & Blog network & Node \\
USA-Airport~\cite{ribeiro2017struc2vec} & 1190 & 13599 & 4 & 1 & A. & Cov.\&Con. & Country &  Air-traffic network & Node \\
Brazil-Airport~\cite{ribeiro2017struc2vec} & 131 & 1038 & 4 & 1 & A. & Cov.\&Con. & Country& Air-traffic network & Node  \\
Europe-Airport~\cite{ribeiro2017struc2vec} & 399 & 5999 & 4 & 1 & A. & Cov.\&Con. & Country & Air-traffic network & Node  \\ 
Yeast~\cite{morris2020tudataset} & 21.54 & 22.84 & 2 & 79601 & A. & Cov. & Species & Protein network & Node/Edge \\
OGBL-PPA~\cite{hu2020open} & 576289 & 30326273 & - & 1 & G. & Cov. & Throughput & Protein network & Edge   \\
OGBL-COLLAB~\cite{hu2020open} & 235868 & 1285465 & - & 1  & G. & Con. & Time & Collaboration network & Edge \\
\bottomrule
\end{tabular}}
\label{tab:datasets}
\end{table*}

Evaluating models under distribution shifts in graph learning is crucial for assessing their generalizability and adaptability. In this section, we introduce common datasets and benchmarks that are typically used for such tasks, covering a range of fields and types of distribution shifts.

\smallskip 
\noindent\textbf{Open Graph Benchmark (OGB)~\cite{hu2020open}.} OGB is a widely used platform that provides a collection of standardized datasets for machine learning on graphs. It includes diverse task types such as node classification, graph classification, and link prediction, covering large-scale molecular graphs, biological networks, and social networks. OGB datasets are often used to study how models generalize under distribution changes or domain shift scenarios. Notable datasets such as OGBG-MolPCBA and OGBG-PPA provide molecular and protein interaction networks, respectively, and are commonly utilized for OOD generalization and adaptation tasks.

\smallskip 
\noindent\textbf{TUdataset~\cite{morris2020tudataset}.} TUdataset is another popular collection of benchmark datasets, including graph classification and regression tasks across different domains, such as bioinformatics (NCI1, D\&D) and social networks (IMDB, Reddit). These datasets are frequently used to evaluate model generalization across graphs of varying sizes and types, making them useful for cross-domain tasks that involve distribution shifts in graph structures and attributes.

\smallskip 
\noindent\textbf{GOOD Benchmark (Graph Out-of-Distribution Benchmark)~\cite{gui2022good}.}
The GOOD benchmark is specifically designed to assess the performance of graph learning models under out-of-distribution (OOD) scenarios. GOOD includes several tasks where models need to handle shifts in graph data distribution, such as covariate and concept shifts. For example, the Spurious-Motif dataset in GOOD challenges models to discern between spurious and true features within graphs, offering a robust test for generalization under distribution shifts.

\smallskip 
\noindent\textbf{DrugOOD~\cite{ji2022drugood}.}
DrugOOD addresses OOD challenges in AI-aided drug discovery, particularly affinity prediction tasks with noisy annotations. This benchmark includes molecular graph datasets and evaluates model performance across various settings, such as Assay, Scaffold, and Size splits, where domain-specific knowledge is crucial for generalization under distribution shifts. DrugOOD primarily considers covariate shifts, thus providing insights into domain generalization for molecular graphs.

Notably, most existing graph learning benchmarks include only one type of shift.
In addition to the benchmarks mentioned above, other widely-used datasets in graph learning tasks under distribution shifts include citation networks like Cora, Citeseer, and Pubmed, which are commonly used for node classification tasks. 
Table~\ref{tab:datasets} summarizes the commonly used datasets for graph learning under distribution shifts, detailing their properties such as the average number of nodes, edges, classes, and graphs, as well as the applicable task types and distribution shift scenarios. These datasets cover the diversity of challenges faced in graph learning and highlights the importance of evaluating models across multiple types of distribution shifts.

\section{Future Directions}
\label{sec:future}
\textbf{Theoretical Understanding of Graph Distribution Shifts.}
Future theoretical analyses could delve deeper into the feasibility and effectiveness of graph OOD adaptation, particularly in scenarios where the label function changes between training and test data.
There is also a need to develop theories and methodologies specifically tailored for graph data or graph models, taking into account the intricate structural information inherent in graphs~\cite{wu2024distributional}.
Furthermore, it is worth exploring more diverse scenarios, such as universal domain adaptive node classification \cite{Chen2023UniversalDA}, graph size adaptation \cite{yehudai2021local}, and multi-source transfer.
Notably, several generalization bounds are derived from the graph transferability evaluation perspective 
 \cite{Ruiz2020GraphonNN}, \cite{Zhu2020TransferLO}, \cite{Chuang2022TreeMD}, and may 
assist in selecting high-quality source graphs in the multi-source transfer setting. 
However, identifying the optimal combination of source graphs with theoretical guarantees remains an open problem.

\smallskip
\noindent
\textbf{Label-efficient Test-time Graph OOD Adaptation.}
Test-time adaptation has garnered increasing attention in traditional machine learning, yet relatively few works have been conducted for graph settings. One major challenge is performing test-time adaptation effectively with limited labeled data. When only a few labeled instances are available, there can be significant distribution shifts between the labeled data and and the vast amount of unlabeled test data. With sparse labeled examples, the adaptation process may struggle to accurately reflect the underlying distribution of the target domain. At the same time, if the labeled data do not adequately represent the test distribution, the model's ability to generalize effectively from these limited labels is compromised. To address these challenges, future research could explore techniques such as semi-supervised learning, active learning, and self-training approaches specifically tailored for graph data.

\smallskip
\noindent
\textbf{Trustworthy Graph Learning under Distribution Shifts.} Although research in trustworthy graph learning (e.g., fairness~\cite{dong2023fairness}, explainability~\cite{yuan2023explainability}, reliability~\cite{shi2023calibrate}) has been extensively studied in recent years, tackling distribution shifts in trustworthy graph learning remains under-explored. Current notations, methodologies, and metrics on trustworthiness may not be directly applicable to scenarios with distribution shifts. For instance, a fair graph learning model might not achieve the same level of fairness when generalizing it to an unseen graph. This calls for the development of new theories and models specifically designed for trustworthy graph learning in OOD and adaptation scenarios. Addressing these gaps could significantly advance reliable graph-based decision-making in dynamic and shifting environments.

\smallskip
\noindent
\textbf{Distribution Shifts on Complex Graphs.}
In contrast to the substantial efforts dedicated to addressing distribution shifts on regular graphs, studies on more complex graph types, such as 
spatial-temporal, heterogeneous graphs, have received comparatively less attention~\cite{hu2024prompt}. Such complex graphs often exhibit diverse and dynamic patterns or involve entities and relationships of various types, introducing more intricate and nuanced distribution shifts. 
The comprehensive exploration and efficient mitigation of distribution shifts on complex graph types are pivotal for enhancing the capabilities of graph machine learning in broader scenarios, such as recommendation systems, healthcare systems, and traffic forecasting.

\smallskip
\noindent
\textbf{Graph Foundation Model for Combating Distribution Shift.}
Developing foundation models specifically tailored for graphs holds significant potential for addressing distribution shifts. 
These models could integrate comprehensive graph representation learning with adaptation strategies to enhance generalization across varying domains~\cite{xia2024anygraph}. From the algorithm perspective, developing graph foundation models involves creating architectures that are flexible and adaptive to distribution shifts with minimal fine-tuning. This allows for efficient adaptation to new distributions while preserving the integrity of learned representations.
Furthermore, efficient fine-tuning methods, such as graph prompt tuning, are essential for maintaining the generalization capabilities of these models in dynamic, real-world applications~\cite{zhao2024all}. Such techniques can ensure that the models remain adaptable and effective when faced with shifts in graph topology and attributes.

\section{Conclusion}
\label{sec:conclusion}
In this survey, we comprehensively analyze existing methods related to deep graph learning under distribution shifts, covering three critical scenarios: graph OOD generalization, training-time graph OOD adaptation, and test-time graph OOD adaptation.
First, we establish problem definitions and explore different graph distribution shift types that can occur.
Following this, we discuss related topics and explain our categorization.
Based on the proposed taxonomy, 
we systematically review techniques aimed at enhancing model generalizability, as well as methods for training-time and test-time adaptation
exploring both model-based and data-centric approaches.
Furthermore, we provide a summary of the datasets commonly utilized in this research area, highlighting their significance and relevance to the challenges associated with distribution shifts.
Finally, we emphasize several challenges that remain and outline promising future research directions. 
This survey aims to serve as a valuable resource for researchers looking to gain a deeper understanding of the progress and emerging trends in this rapidly evolving area of study.

\bibliographystyle{IEEEtran}
\bibliography{reference}

\vfill

\end{document}